\documentclass[10pt,a4paper]{article}
\usepackage[cmex10]{amsmath}
\usepackage{amssymb}
\usepackage{amsthm}
\usepackage{bm}
\usepackage{algorithm}
\usepackage[noend]{algpseudocode}
\usepackage{color}
\usepackage{caption}
\usepackage{subcaption}
\usepackage{graphicx}
\usepackage{url}

\DeclareMathOperator{\argmax}{arg\,max}

\newcommand{\Real}{\mathbb{R}}
\newcommand{\Nat}{\mathbb{N}}

\makeatletter
\newcommand{\SPAN}{\mathop{\operator@font span}}
\newcommand{\sinc}{\mathop{\operator@font sinc}}
\newcommand{\RANK}{\mathop{\operator@font rank}}
\newcommand{\bigO}{\mathrm O}
\makeatother
\makeatletter
\newcommand{\supp}{\mathop{\operator@font supp}}
\newcommand{\rank}{\mathop{\operator@font rank}}
\newcommand{\diag}{\mathop{\operator@font diag}}
\makeatother
\makeatletter
\def\BState{\State\hskip-\ALG@thistlm}
\makeatother

\newcommand{\ubar}[1]{\text{\b{$#1$}}}
\newcommand{\orf}{\ubar{f}}
\newcommand{\orfb}{\ubar{\bm{f}}}
\newcommand{\oru}{\ubar{u}}
\newcommand{\orub}{\ubar{\bm{u}}}
\newcommand{\orc}{\ubar{c}}
\newcommand{\ora}{\ubar{\alpha}}
\newcommand{\orab}{\ubar{\bm{\alpha}}}
\newcommand{\orutb}{\ubar{\bm{\theta}}}
\newcommand{\org}{\ubar{g}}

\newtheorem{theorem}{Theorem}

\newtheorem{remark}{Remark}

\usepackage{array}
\makeatletter
\newcommand{\thickhline}{%
    \noalign {\ifnum 0=`}\fi \hrule height 1pt
    \futurelet \reserved@a \@xhline
}
\newcolumntype{"}{@{\hskip\tabcolsep\vrule width 1pt\hskip\tabcolsep}}
\makeatother

\begin{document}

\title{Robust Non-linear Regression: A Greedy Approach Employing Kernels with Application to Image Denoising}

\author{George~Papageorgiou\footnote{geopapag@di.uoa.gr} , Pantelis~Bouboulis\footnote{panbouboulis@gmail.com} and~Sergios~Theodoridis\footnote{stheodor@di.uoa.gr}
}
\maketitle

\begin{abstract}
We consider the task of robust non-linear regression in the presence of both inlier noise and outliers. Assuming that the unknown non-linear function belongs to a Reproducing Kernel Hilbert Space (RKHS), our goal is to estimate the set of the associated unknown parameters. Due to the presence of outliers, common techniques such as the Kernel Ridge Regression (KRR) or the Support Vector Regression (SVR) turn out to be inadequate. Instead, we employ sparse modeling arguments to explicitly model and estimate the outliers, adopting a greedy approach. The proposed robust scheme, i.e., Kernel Greedy Algorithm for Robust Denoising (KGARD), is inspired by the classical Orthogonal Matching Pursuit (OMP) algorithm. Specifically, the proposed method alternates between a KRR task and an OMP-like selection step. Theoretical results concerning the identification of the outliers are provided. Moreover, KGARD is compared against other cutting edge methods, where its performance is evaluated via a set of experiments with various types of noise. Finally, the proposed robust estimation framework is applied to the task of image denoising, and its enhanced performance in the presence of outliers is demonstrated.
\end{abstract}

\section{Introduction}
\label{sec:intro&prob_statem}

%
%
%
%
The problem of function estimation has attracted significant attention in the machine learning and signal processing communities over the past decades. In this paper, we target the specific task of regression, which is typically described as follows: given a training set of the form $\mathcal{D}=\left\{ \left( y_i, \bm{x}_i \right) \right\}_{i=1}^N$, we aim to estimate the input-output relation between $\bm{x}_i$ and $y_i$; i.e., a function $f$, such that $f(\bm{x}_i)$ is ``close'' to $y_i$, for all $i$. This is usually achieved by employing a \textit{loss function}, i.e., a function $C(\bm{x}_i, y_i, f(\bm{x}_i))$,  that measures the deviation between the observed values, $y_i$, and the predicted values, $f(\bm{x}_i)$, and minimizes the so called \textit{Empirical Risk}, i.e., $\sum_{i=1}^N C(\bm{x}_i, y_i, f(\bm{x}_i))$. For example, in the least squares regression, one adopts the squared error, i.e., $\left( y_i - f(\bm{x}_i) \right)^2$, which leads to the minimization of a quadratic function.

Naturally, the choice for $f$ strongly depends on the underlying nature of the data. In this paper, we assume that $f$ belongs to an RKHS. These are inner product function spaces, in which every function is reproduced by an associated (space defining) kernel; that is, for every $\bm{x} \in \mathcal{X}$, there exists $\kappa(\cdot,\bm{x})\in \mathcal{H}$, such that $f(\bm{x})= \langle f, \kappa (\cdot, \bm{x})\rangle_{\mathcal{H}}$. This is the case that has been addressed (amongst others) by two very popular and well-established methods which are commonly referred to as the \textit{Kernel Ridge Regression} (KRR) and the \textit{Support Vector Regression} (SVR).

Another important issue that determines the quality of the estimation is the underlying noise model. For the most common noise sources (Gaussian, Laplacian, etc.) the estimation is performed via the KRR by solving a (regularized) Least Squares task, \cite{theodoridis2015machinelearning}. However, when outliers are present or when the noise distribution exhibits long tails (commonly originating from another noisy source) the performance of the KRR degrades significantly. The sensitivity of the Least Squares estimator to outliers is well known and studied, even for the simplest case of the linear regression task; a variety of methods, that deal with this problem, have been established over the years, e.g., \cite{huber1981wiley, maronna2006robust, rousseeuw2005robust, huber19721972, rousseeuw1990unmasking, leroy1987robust, razavi2012robust, boyd2011distributed, boyd2011alternating, wipf2004sparse, jin2010algorithms, theodoridis2015machinelearning, papageorgiou2014robust}. On the other hand, the development of robust estimators for the KRR has been addressed only recently; the task is known as the Robust Kernel Ridge Regression (RKRR), \cite{mateos2012robust,mitra2010robust}. In this case, $y_i$ is assumed to be generated by
\begin{equation}
\label{eq:nonl_equation_inlier_only}
y_i = \orf(\bm{x}_i)+v_i,\ i=1,...,N,
\end{equation}
where $v_i$ are random noise samples which may contain outliers. The present paper focuses on this task in the special case where the unknown function, $\orf$ (the underbar denotes the original function that we wish to estimate), is assumed to lie in an RKHS, $\mathcal{H}$. It should be noted that both SVR and KRR can be employed to address this problem, but the presence of outliers degrades their performance significantly due to over-fitting, \cite{ROKS, bouboulis2010adaptive}. Of course, in SVR this effect is not as dominant, due to the $\ell_1$ loss function that it is employed, in contrast to the typical KRR, which is the subject of this work.

So far, there exist two paths for addressing the task of non-linear regression in RKH spaces, which involve sparse modeling techniques to cope with robustness against the presence of outliers: a) the regularized by the $\ell_1$-norm squared error minimization method (\cite{mateos2012robust}) and b) a sparse Bayesian learning approach (\cite{mitra2010robust,tipping2001sparse}). The first method, which has been used for load curve data cleansing, identifies the outliers (modeled as a sparse vector) by employing the $\ell_1$-norm, while the sparsity level is controlled by the tuning of a regularization parameter. On the other hand, the second method introduces Bayesian techniques via the use of hyper-parameters in order to infer the outlier estimates. It should be noted that, the authors of the first method also proposed a refinement of their technique by employing weights on the $\ell_1$-norm, iteratively; this resulted to an enhanced performance (in terms of the estimation error). The specific weighted regularization, which is based on a function of the outliers used as weights, it was shown to approximate the $\ell_0$-norm of the sparse outlier vector. Despite the overall good performance and low computational requirements of the refined $\ell_1$-norm method, its major drawback is the requirement for fine tuning of two regularization parameters; in practice, this can be quite demanding process that can limit its potential in real life applications. On the other hand, the Bayesian model does not require any kind of parameter tuning; however, its computational cost can be discouraging. For example, the typical time required for the denoising of an image is significantly higher, compared to the newly proposed method, as it is discussed in the experimental section of the paper.

Our proposed scheme manages to efficiently balance between the best aspects of the aforementioned methods in terms of: a) estimation performance, b) computational efficiency and c) simplicity, since automatic parameter tuning is also established for the task of image denoising. Moreover, theoretical properties regarding the identification of the outliers have been established. It should be noted that, this is a result that has been presented for the first time in the related bibliography. The proposed method adopts a model of the form $y=f(\bm{x})$, where $f \in\mathcal{H}$, and a decomposition of the noise into two parts, a sparse outlier vector $\bm{u}$ and the inlier vector $\bm{\eta}$. Next, a two step algorithmic procedure is employed, attempting to estimate both the outliers and also the original (unknown) function $\orf$. This is accomplished by alternating between a) a greedy-type algorithm based on the popular \textit{Orthogonal Matching Pursuit} (OMP) \cite{pati1993orthogonal, tropp2004greed, tropp2007signal}, that selects the dominant outlier sample in each step, and b) a kernel ridge regression task in order to update the current non-linear estimate. Results regarding the identification of the outliers are also provided. Moreover, comparisons against the previously published approaches, based on the Bayesian framework and on the minimization of the $\ell_1$-norm for the sparse outlier vector, are performed. Of course, the application areas of the robust regression task are not only limited to the image denoising task, but also include geostatistics, medical statistics, etc.

The rest of the paper is organized as follows. In Section \ref{sec:prelimin}, the basic properties of RKHS are summarized, and in Section \ref{sec:relat_work} the problem is formulated and related state-of-the-art methods are presented. Next, in Section \ref{sec:proposed_algorithm}, the proposed scheme is introduced and described in detail. Section \ref{sec:theoretical_anal} provides the theoretical results regarding the identification of the outliers. In Section \ref{sec:exper}, extended tests against other cutting edge methods are performed. The efficiency of each method is depicted in terms of: a) the achieved mean square error (MSE) and b) the convergence time. In Section \ref{sec:image_den}, the method is applied to the task of robust image denoising in order to remove the noise component that comprises a mix of impulsive and Gaussian sources. Finally, Section \ref{sec:concl} is devoted to the summary of this work.  

\textbf{Notation}: Throughout this work, capital calligraphic letters are employed to denote sets, e.g., $\mathcal{S}$, where $\mathcal{S}^{c}$ denotes the complement of $\mathcal{S}$. Function are denoted by small letters, e.g., $f$; in particular, the underbar denotes the unknown function that we wish to ``learn", i.e., $\orf$, while the hat denotes the respective estimate, i.e., $\hat{f}$. Small letters denote scalars, e.g., $\varepsilon$, while bold capital letters denote matrices, e.g., $\bm{X}$, bold lowercase letters are reserved for vectors, e.g., $\bm{\theta}$ (each vector is regarded as a column vector) and the symbol $\cdot^T$ denotes the transpose of the respective matrix/vector. Also, $\diag(\bm{a})$, where $\bm{a}$ is a vector, denotes the respective square diagonal matrix\footnote{This matrix has the vector's coefficients on its \textit{diagonal}, while all other entries are equal to zero.}, while $\supp(\bm{a})$ denotes the support set of the vector $\bm{a}$. The $j$-th column of matrix $\bm{X}$ is denoted by $\bm{x}_j$ and the element of the $i$-th row and $j$-th column of matrix $\bm{X}$ by $x_{ij}$. Moreover, the $i$-th element of vector $\bm{\theta}$ is denoted by $\theta_i$. An arithmetic index in parenthesis, i.e., $(k)$, $k=0,1,\dots$, is reserved to declare an iterative (algorithmic) process, e.g., on matrix $\bm{X}$ and vector $\bm{r}$ the iteratively generated matrix and vector are denoted by $\bm{X}_{(k)}$ and $\bm{r}_{(k)}$, respectively. Following this rationale, $r_{(k),i}$ is reserved for the $i-$th element of the iteratively generated vector $\bm{r}_{(k)}$. The notation $\bm{X}_{\mathcal{S}}$ denotes the columns of matrix $\bm{X}$ restricted over the set $\mathcal{S}$, while $\bm{X}_{\mathcal{S},\mathcal{S}}$ denotes the restriction of his rows and columns over $\mathcal{S}$. Accordingly, the notation $\bm{u}_{\mathcal{S}}$ denotes the elements of vector $\bm{u}$, restricted over the set $\mathcal{S} \subseteq \supp( \bm{u})$. Finally, the identity matrix of dimension $N$ will be denoted as $\bm{I}_N$, where $\bm{e}_j$ is its $j$-th column vector, the zero matrix of dimension $N\times N$, as $\bm{O}_{N},$ the vector of zero elements of appropriate dimension as $\bm{0}$ and the columns of matrix $\bm{I}_N$ restricted over the set $\mathcal{S}$, as $\bm{I}_{\mathcal{S}}$.

\section{Preliminaries}
\label{sec:prelimin}
\label{ssec:kernel_theory}

In this section, an overview of some of the basic properties of the RKHS is provided \cite{theodoridis2015machinelearning, smolalearningwithKer, Cristianini, elsevier_kernels, theodoridis2008koutroumbas,Aronszajn_1950_9268}.
An RKHS is a Hilbert space $\mathcal{H}$ over a field $\mathbb{F}$ for which there exists a positive definite function, $\kappa:\mathcal{X}\times \mathcal{X}\rightarrow\mathbb{F}$, such that for every $\bm{x}\in \mathcal{X}$, $\kappa(\cdot,\bm{x})$ belongs to $\mathcal{H}$ and $f(\bm{x})=\langle f,\kappa(\cdot, \bm{x})\rangle_{\mathcal{H}}, \textrm{ for all } f\in\mathcal{H}$; in particular, $\kappa(\bm{x},\bm{y})=\langle \kappa(\cdot, \bm{y}), \kappa(\cdot, \bm{x})\rangle_{\mathcal{H}}$.
The \textit{Gram} matrix $\bm{K}$, corresponding to the kernel $\kappa$, i.e., the matrix with elements $\kappa_{ij} := \kappa(\bm{x}_i, \bm{x}_j)$, is positive (semi) definite for any selection of a finite number of points $\bm{x}_1, \bm{x}_2,\dots, \bm{x}_N$, $N \in \Nat^*$. Moreover, the fundamental Representer Theorem establishes that although an RKHS may have infinite dimension, the solution of any regularized regression optimization task lies in the span of $N$ specific kernels, e.g., \cite{smolalearningwithKer, theodoridis2015machinelearning}. In other words, each minimizer $f \in \mathcal{H}$ admits a representation of the form $f=\sum_{j=1}^N \alpha_j \kappa (\cdot, \bm{x}_j)$. However, in many applications (also considered here) a \textit{bias} term, $c$, is often included in the aforementioned expansion; i.e., we assume that $f$ admits the following representation:
\begin{equation}
\label{eq:bias_repres}
f = \sum_{j=1}^N \alpha_j \kappa(\cdot, \bm{x}_j) + c.
\end{equation}
The use of the bias term is theoretically justified by the Semi-parametric Representer Theorem, e.g., \cite{smolalearningwithKer, theodoridis2015machinelearning}.

Although there are many kernels to choose from, throughout this manuscript we have used the real \textit{Gaussian radial basis function} (RBF), i.e., $\kappa_{\sigma}(\bm{x},\bm{x}') : = \exp\left(-\|\bm{x}-\bm{x}'\|^2/\sigma^2\right)$, defined for $\bm{x}, \bm{x}' \in \mathbb{R}^M$,
where $\sigma$, is a free positive parameter that defines the shape of the kernel function. In the following, $\kappa$ is adopted to denote the Gaussian RBF. An important property of this kernel is that the corresponding matrix, $\bm{K}$, given by $\kappa_{ij}:=\exp(-\frac{||\bm{x}_i-\bm{x}_j ||^2}{\sigma^2}),$
has full rank. The significance of the theorem is that the points $\kappa(\cdot,\bm{x}_1),\kappa(\cdot,\bm{x}_2),...,\kappa(\cdot,\bm{x}_N) \in \mathcal{H}$ are linearly independent, i.e., span the $N$-dimensional subspace of $\mathcal{H}$, \cite{smolalearningwithKer}.

\section{Problem Formulation and Related Works}
\label{sec:relat_work}

\subsection{Robust Ridge Regression in RKHS}
\label{ssec:formulation}

Given the data set  $\mathcal{D}=\left\{ \left(y_i, \bm{x}_i \right)\right\}_{i=1}^N$, we assume that each observation $y_i$ is related to the corresponding input vector, $\bm{x}_i$, via
\begin{align}
y_i = \orf(\bm{x}_i) + \oru_i + \eta_i,\ i=1,\dots,N,
\label{eq:model_ori}
\end{align}
where $\orf \in \mathcal{H}$ and $\mathcal{H}$ is a specific RKHS. The variable $\oru_i$ represents a possible outlier sample and $\eta_i$ a noise component. In a more compact form, this can be cast as $\bm{y} = \orfb  + \orub + \bm{\eta}$, where $\orfb$ is the vector containing the values $\orf(\bm{x}_i)$ for all $i=1,\dots,N$. As $\orub$ represents the vector of the (unknown) outliers, it is reasonable to assume that this is a sparse vector. Our goal is to estimate the input-output relation $\orf$ from the noisy observations of the data set $\mathcal{D}$. This can be interpreted as the task of simultaneously estimating both a sparse vector $\bm{u}$ and as well as a function $f\in\mathcal{H}$, that maintains a low squared error for $L(\mathcal{D},f,\bm{u})=\sum_{i=1}^N \left( y_i - f(\bm{x}_i) - u_i \right)^2$. Moreover, motivated by the representer theorem, we adopt the representation in \eqref{eq:bias_repres}, as a means to represent the solution for $f$.  Under these assumptions, equation \eqref{eq:model_ori} can be expressed in a compact form as
\begin{align}
\bm{y} = \bm{K} \orab +\orc \bm{1} + \orub +  \bm{\eta} = \bm{X}_{(0)} \begin{pmatrix}
\orab\\ \orc
\end{pmatrix}  + \bm{v},
\label{eq:model_ori_compact}
\end{align}
where $\bm{K}$ is the kernel matrix, $\bm{X}_{(0)} =[\bm{K}\ \bm{1}]$ ($\bm{1}$ is the vector of ones) and $\bm{v}= \orub + \bm{\eta}$ is the total noise vector (outlier plus inlier). Accordingly, the squared error is written as
$L(\mathcal{D},\bm{\alpha},c,\bm{u})=\|\bm{y} - \bm{K}\bm{\alpha} - c\bm{1} - \bm{u}\|_2^2,$ and we cast the respective minimization task as:
\begin{align}
\begin{matrix}
\displaystyle{\min_{\bm{u},\bm{\alpha}\in\mathbb{R}^N, c\in\mathbb{R}}}  & \|\bm{u}\|_0\\
\textrm{s. t.}  &  \|\bm{y} - \bm{K}\bm{\alpha} - c\bm{1} - \bm{u}\|^2_2 + \lambda \left\| \begin{pmatrix} \bm{\alpha} \\ c \end{pmatrix}  \right\|_2^2 \leq \varepsilon,
\end{matrix}
\label{eq:model_opt}
\end{align}
for some predefined parameters $\varepsilon, \lambda>0$, where we have also used a standard regularization term (via $\lambda$) in order to keep the norm of the vector for the kernel expansion coefficients low. An alternative regularization strategy, which is common in the respective literature (based on KRR), is to include the norm of $f$, i.e., $\|f\|_{\mathcal{H}}^2 = \bm{\alpha}^T\bm{K}\bm{\alpha}$, instead of the norm of the coefficients' vector, leading to the following task:
\begin{align}
\begin{matrix}
\displaystyle{\min_{\bm{u},\bm{\alpha}\in\mathbb{R}^N, c\in\mathbb{R}}}  & \|\bm{u}\|_0\\
\textrm{s. t.}  &  \|\bm{y} - \bm{K}\bm{\alpha} - c\bm{1} - \bm{u}\|^2_2 + \lambda\bm{\alpha}^T\bm{K}\bm{\alpha} \leq \varepsilon.
\end{matrix}
\label{eq:model_opt2}
\end{align}

\subsection{Related Works}
\label{ssec:related works}

As already discussed, two methods that deal with the RKRR task have been previously established. The first method is based on the minimization of the regularized cost via the $\ell_1$-norm (or of a variant of it) of the sparse outlier vector (instead of minimizing the respected $\ell_0$-norm) and the second one employs sparse Bayesian learning arguments.\\

\noindent\textbf{RAM: Refined Alternating Directions Method of Multipliers}\\
The method is based on the $\ell_1$-norm regularized minimization for the sparse outlier vector (similar to the formulation in \eqref{eq:model_opt2}). By replacing the $\ell_0$-norm with its closest convex relaxation, i.e., the $\ell_1$-norm, we resort to solving a convex optimization task instead. The authors have established the so-called AM solver, which employs the alternating directions method of multipliers (ADMM) in order to solve the task in its LASSO formulation. Furthermore, having obtained this solution as an initialization, they have improved the scheme via the use of the reweighted $\ell_1$-norm technique, as proposed in \cite{candes2008enhancing}. The resulting method is called RAM and stands for refined AM solver. More details over the scheme can be found in \cite{mateos2012robust}.\\
\noindent\textbf{RB-RVM: Robust Relevance Vector Machine - Sparse Bayesian Learning}\\
The Sparse Bayesian learning scheme is based on the RVM rationale and it employees hyper-parameters in order to infer not only the unknown kernel coefficients but also the sparse outlier estimate. More details on this approach can be found in \cite{mitra2010robust,tipping2001sparse}.

\section{Kernel Greedy Algorithm for Robust Denoising (KGARD)}
\label{sec:proposed_algorithm}

\subsection{Motivation and Proposed Scheme}
\label{subsec:motiv}

In the following, we build upon the two formulations \eqref{eq:model_opt} and \eqref{eq:model_opt2}, that attempt (and indeed succeed) to solve the robust Least Squares task via the use of a scheme inspired by the basic greedy algorithm, i.e., the OMP. Obviously, their difference lies solely on the regularization term. In the first approach, the regularization is performed using the $\ell_2$-norm of the unknown kernel parameters (which is a standard regularization technique in linear methods). In the alternative formulation, i.e., \eqref{eq:model_opt2}, we perform the regularization via the $\mathcal{H}$-norm of $f$. Although we have extensively tested both methods, we have noticed that \eqref{eq:model_opt} leads to improved performance. Thus, we have presented the method in a general form and depending on the selection of a matrix, each one of the two tasks (\eqref{eq:model_opt} or \eqref{eq:model_opt2}) can be solved.

Since both tasks in \eqref{eq:model_opt} and \eqref{eq:model_opt2} are known to be NP-hard, a straight-forward computation of a solution seems impossible. However, under certain assumptions, greedy-based techniques often manage to provide accurate solutions to $\ell_0$-norm minimization tasks, which are also guaranteed to be close to the optimal solution. The proposed Kernel Greedy  Algorithm for Robust Denoising (KGARD), which is based on a modification of the popular Orthogonal Matching Pursuit (OMP), has been adapted to both formulations, i.e., \eqref{eq:model_opt} and \eqref{eq:model_opt2}, as presented in Algorithm \ref{alg:RGARD}.

First, one should notice that, the quadratic inequality constraint could also be written in a more compact form as follows:
\begin{flalign}
&& J(\bm{z})=\| \bm{y} - \bm{Xz}\|_{2}^2 &+ \lambda \bm{z}^{T}\bm{B} \bm{z} \leq \varepsilon,& \label{eq:J_def} \\
\text{where}&& \bm{X}=\begin{bmatrix}
\bm{K} & \bm{1} & \bm{I}_N
\end{bmatrix},&\ \bm{z}= ( \bm{\alpha}^T , c , \bm{u}^T )^T,&
\label{eq:X_matrix}
\end{flalign}
and for the choice of matrix $\bm{B}$ either one of the following matrices can be used,
\begin{equation}
\bm{B}=\begin{bmatrix} \bm{I}_N & \bm{0} & \bm{O}_N \\ \bm{0}^{T} & 1 & \bm{0}^{T}\\ \bm{O}_N & \bm{0} & \bm{O}_N
\end{bmatrix}\ \text{or}\ \begin{bmatrix} \bm{K} & \bm{0} & \bm{O}_N \\ \bm{0}^{T} & 0 & \bm{0}^{T}\\ \bm{O}_N & \bm{0} & \bm{O}_N
\end{bmatrix},
\label{eq:Bmatrix}
\end{equation}
depending on whether \eqref{eq:model_opt} or \eqref{eq:model_opt2} is adopted, respectively. 

\begin{algorithm}[t]
\caption{Kernel Greedy Algorithm for Robust Denoising: KGARD}
\label{alg:RGARD}
\begin{algorithmic}[1]
\Procedure{KGARD}{$\bm{K},\ \bm{y},\ \lambda,\ \epsilon$}
\State $k\gets 0$
\State $\widetilde{\mathcal{S}}_{0} \gets \{1,2,...,N+1 \},\  \mathcal{S}_{0}^c \gets \{N+2,...,2N+1 \}$, $\bm{X}=[\bm{K}\ \bm{1}\ \bm{I}_N],\ \bm{B}$ in \eqref{eq:Bmatrix}
\State  $\hat{\bm{z}}_{(0)} \gets \left(\bm{X}_{\widetilde{\mathcal{S}}_{0}}^T\bm{X}_{\widetilde{\mathcal{S}}_{0}}+\lambda \bm{B}_{\widetilde{\mathcal{S}}_{0}, \widetilde{\mathcal{S}}_{0}} \right)^{-1}\bm{X}_{\widetilde{\mathcal{S}}_{0}}^T \bm{y}$ 
\State  $\bm{r}_{(0)} \gets \bm{y} - \bm{X}_{\widetilde{\mathcal{S}}_{0}} \hat{\bm{z}}_{(0)}$
\While{$\| \bm{r}_{(k)} \|_2 > \epsilon$}
\State $k \gets k+1$
\State $j_k \gets \argmax_{j \in \mathcal{S}_{k-1}^c} |r_{(k-1),j}|,\ i_k=j_k+|\widetilde{\mathcal{S}}_0|$
\State $\widetilde{\mathcal{S}}_{k} \gets \widetilde{\mathcal{S}}_{k-1} \cup \{i_k\}  ,\ \mathcal{S}_{k}^c \gets \mathcal{S}_{k-1}^c \setminus \{j_k\}$
\State $\hat{\bm{z}}_{(k)} \gets \left(\bm{X}_{\widetilde{\mathcal{S}}_{k}}^T\bm{X}_{\widetilde{\mathcal{S}}_{k}}+\lambda \bm{B}_{\widetilde{\mathcal{S}}_{k}, \widetilde{\mathcal{S}}_{k}} \right)^{-1}\bm{X}_{\widetilde{\mathcal{S}}_{k}}^T \bm{y} $ 
\State $\bm{r}_{(k)} \gets \bm{y} - \bm{X}_{\widetilde{\mathcal{S}}_{k}} \hat{\bm{z}}_{(k)} $
\EndWhile
\State \textbf{Output:} $\hat{\bm{z}}_{(k)}=\left( \hat{\bm{\alpha}}_{(k)}^T,\hat{c}_{(k)},\hat{\bm{u}}_{(k)}^T \right)^T$ after $k$ iterations.
\EndProcedure
\end{algorithmic}
\end{algorithm}

The proposed method, as presented in Algorithm \ref{alg:RGARD}, attempts to solve the task \eqref{eq:model_opt} or \eqref{eq:model_opt2}, via a sparse greedy-based approach. The algorithm alternates between an LS task and a column selection step, that enlarges the solution subspace at each step, in order to minimize the residual error. The scheme shares resemblances to the OMP algorithm. Its main differences, are: (a) the solution of a regularized LS task at each iteration (instead of a simple LS task), i.e.,
\begin{align}
\min_{\bm{z}} J_k(\bm{z}) = \min_{\bm{z}} \left\{\| \bm{y} - \bm{X}_{\mathcal{S}_{k}}\bm{z}\|_{2}^2+ \lambda \bm{z}^{T}\bm{B}_{\widetilde{\mathcal{S}}_{k}, \widetilde{\mathcal{S}}_{k}} \bm{z}\right\} ,  \label{eq:J_opt}
\end{align}
and (b) the use of a specific initialization on the solution and the residual. These seemingly small differences dictate for a completely distinct performance analysis for the method  as compared to the standard OMP. The method is described best, via the use of subsets, corresponding to a set of \textit{active} and \textit{inactive} columns for matrix $\bm{X}$. The active set, $\widetilde{\mathcal{S}}_k$, which includes indices of the active columns\index{Active columns} from $\bm{X}$ at the $k$-th step, and the inactive set\index{Inactive set}, $\widetilde{\mathcal{S}}_k^c$, which contains the remaining ones, i.e., those that do not participate in the representation. Moreover, the set of indices that refer to the selected columns of the identity matrix, $\bm{I}_N$, and with respect to the set $\widetilde{\mathcal{S}}$, is defined as:
\begin{align}
\mathcal{S}_k:= \left\{ \begin{array}{cl}
\left\{ j-|\widetilde{\mathcal{S}}_0|\ :\ j \in \widetilde{\mathcal{S}}_k \setminus \widetilde{\mathcal{S}}_0 \right\}\ &\text{for } k =1,2,\dots \\
\emptyset\ &\text{for } k=0
\end{array}
\right.,
\label{eq:proj_set_indices}
\end{align}
where $|\widetilde{\mathcal{S}}_0|$ denotes the cardinality\index{Cardinality} of the set $\widetilde{\mathcal{S}}_0$ and $\widetilde{\mathcal{S}}_k \setminus \widetilde{\mathcal{S}}_0:=\{j:\ j\in \widetilde{\mathcal{S}}_k\ \text{and } j \notin \widetilde{\mathcal{S}}_0\}$. The set $\mathcal{S}_k$ is of major importance, since it indicates the support for the sparse outlier estimate. Also note that, $\mathcal{S}_k^c$ is used for its complementary set. While the set $\widetilde{\mathcal{S}}_{k}^c$ refers to the columns of the augmented matrix $\bm{X}$, the set $\mathcal{S}_{k}^c$ refers to the columns of the identity matrix (the last part of matrix $\bm{X}$), i.e., matrix $\bm{I}_{N}$. Initially, only the first $N+1$ columns of matrices $\bm{X}$ and $\bm{B}$, have been activated. Thus, $k=0$, leads to the initialization of the active set $\widetilde{\mathcal{S}}_0 = \{1,2,\dots, N+1\}$ with the corresponding matrices:
\begin{flalign*}
&& \bm{X}_{\widetilde{\mathcal{S}}_{0}} &= [\bm{K}\ \bm{1}],&\\
 \text{and}\ && \bm{B}_{\widetilde{\mathcal{S}}_{0}, \widetilde{\mathcal{S}}_{0}} = & \bm{I}_{N+1}\
\text{or}\
\begin{bmatrix}
\bm{K} & \bm{0}\\ \bm{0}^T & 0
\end{bmatrix},&
\end{flalign*}
depending on the model selection, i.e., \eqref{eq:model_opt} or \eqref{eq:model_opt2} respectively. Hence, the solution to the initial LS problem, is given by
\begin{align*}
\hat{\bm{z}}_{(0)} \ := \left(\bm{X}_{\widetilde{\mathcal{S}}_{0}}^T\bm{X}_{\widetilde{\mathcal{S}}_{0}} + \lambda\bm{B}_{\widetilde{\mathcal{S}}_{0}, \widetilde{\mathcal{S}}_{0}}\right)^{-1}\bm{X}_{\widetilde{\mathcal{S}}_{0}}^T\bm{y}.
\end{align*}
Next, the method computes the residual $\bm{r}_{(0)} = \bm{y} - \bm{X}_{\widetilde{\mathcal{S}}_0}\hat{\bm{z}}_{(0)}$ and identifies an outlier\footnote{If outliers are not present, the algorithm terminates and no outlier estimate exists in the solution $\hat{\bm{z}}_0$.}, as the largest value of the residual vector. The corresponding index, say $j_1 \in \mathcal{S}_{k-1}^c$, is added into the set of active columns, i.e., $\widetilde{\mathcal{S}}_1 = \widetilde{\mathcal{S}}_0\cup\{i_k\}$. Thus, the matrix $\bm{X}_{\widetilde{\mathcal{S}}_0}$ is augmented by a column drawn from matrix $\bm{I}_{N}$, forming matrix $\bm{X}_{\widetilde{\mathcal{S}}_1}$. Accordingly, the matrix $\bm{B}_{\widetilde{\mathcal{S}}_0,\widetilde{\mathcal{S}}_0}$ is augmented by a zero row and a zero column, forming $\bm{B}_{\widetilde{\mathcal{S}}_1,\widetilde{\mathcal{S}}_1}$. The new LS task is solved again (using matrices $\bm{X}_{\widetilde{\mathcal{S}}_1}$, $\bm{B}_{\widetilde{\mathcal{S}}_1, \widetilde{\mathcal{S}}_1}$) and a new residual $\bm{r}_{(1)}$ is computed. The process is repeated, until the residual drops below a predefined threshold.

Although, both approaches, \eqref{eq:model_opt} and \eqref{eq:model_opt2}, are suitable for dealing with the sparse minimization task, in the experimental set-up used in this paper the selection of \eqref{eq:model_opt} proves a better choice. Henceforth, the model \eqref{eq:model_opt} is adopted. 


\begin{remark}
The matrix $\bm{B}$ in \eqref{eq:Bmatrix} is a projection matrix only for the choice corresponding to the regularization performed with the $\ell_2$-norm (left one). The other choice that depends on the kernel matrix does not have this property. 
\end{remark}

\begin{remark}
In order to simplify the notation, in the next sections, we adopt $\bm{X}_{(k)}$ and $\bm{B}_{(k)}$ to refer to the matrices $\bm{X}_{\mathcal{S}_k}$ and $\bm{B}_{\mathcal{S}_k}$ at the $k$ step.
\end{remark}

\begin{remark}
Once a column has been selected at the $k$-step, it cannot be selected again in any subsequent step, since the corresponding residual coordinate is zero. In other words, the algorithm always selects a column from the last part of $\bm{X}$, i.e., matrix $\bm{I}_N$, that is not included in $\mathcal{S}_{k}.$
\end{remark}

\subsection{Efficient Implementations}
\label{subsec:eff_impl}

As the outliers often comprise a small fraction of the data set, i.e., $k<<N$, a fast implementation time for OMP-like schemes such as KGARD is expected. Initially, the inversion of matrix $\bm{X}_{(0)}^T\bm{X}_{(0)}+\lambda\bm{B}_{(0)}$ plus the multiplication of $\bm{X}_{(0)}^T\bm{y}$, requires $\bigO \left( (N+1)^3 \right)$ flops. At each one of the subsequent steps, the required complexity is $\bigO \left( (N+k+1)^3 \right)$, while the total cost for the method is $\bigO \left( (N+1)^3(k+1)+(5/2)N^2k^2+ (4/3)Nk^3 +k^4/4 \right)$, where $k<<N$. However, the complexity of the method could be further reduced, since a large part of the inverted matrix remains unchanged. To this end, several methods could be employed, \cite{sturm2012comparison}, such as the \textit{matrix inversion lemma} (MIL) or the QR decomposition. However, the most efficient technique proved to be the \textit{Cholesky decomposition} for the matrix to be inverted. The updates are summarized in the following steps:
\begin{itemize}
\item \textit{Replace}  step 4 of algorithm \ref{alg:RGARD}, with:
{\small
\begin{algorithmic}
	\State{Factorization step}: $\bm{M}_{(0)}=\bm{L}_{(0)} \bm{L}_{(0)}^{T}$\\
    \State{Solve} $\bm{L}_{(0)} \bm{L}_{(0)}^{T} \hat{\bm{z}}_{(0)}=\bm{X}_{(0)}^T \bm{y}$ \text{using}:\\
	\begin{itemize}
	\item \text{forward substitution} $\bm{L}_{(0)} \bm{q}=\bm{X}_{(0)}^T\bm{y}$
	\item \text{backward substitution} $\bm{L}_{(0)}^T \hat{\bm{z}}_{(0)} =\bm{q}$
	\end{itemize}
	\State{Complexity}:  $\bigO \left( (N+1)^3/3 + (N+1)^2 \right)$
 \end{algorithmic}
}
\item \textit{Replace} step 10 of algorithm \ref{alg:RGARD}, with:
{\small
\begin{algorithmic}
\State{Compute} $\bm{d}$ \text{such that}: $\bm{L}_{(k-1)}\bm{d}=\bm{X}_{(k-1)}^T \bm{e}_{j_k}$
\State{Compute}: $b=\sqrt{1-||\bm{d}||_2^2}$
\State{Matrix Update}: $ \bm{L}_{(k)}=\begin{bmatrix}
\bm{L}_{(k-1)} & \bm{0}\\ \bm{d}^T & b
\end{bmatrix} $
 \State{Solve}
	$\bm{L}_{(k)} \bm{L}_{(k)}^{T} \hat{\bm{z}}_{(k)} = \bm{X}_{(k)}^T \bm{y}$ \text{using}:\\
	\begin{itemize}
	\item \text{forward substitution} $\bm{L}_{(k)} \bm{p}=\bm{X}_{(k)}^T \bm{y}$
	\item \text{backward substitution} $\bm{L}_{(k)}^T \hat{\bm{z}}_{(k)} =\bm{p}$
	\end{itemize}
	\State{Complexity}:  $\bigO \left( (9/2)N^2 + 5Nk + (3/2)k^2 \right)$ per iteration.
\end{algorithmic}
}
\end{itemize}
Employing the Cholesky decomposition plus the update step leads to a reduction of the total computational cost to $\bigO \left( (N+1)^3/3 + (N+1)^2 + k^3/2 + (5/2)Nk^2  \right)$, which is the fastest implementation for this task (recall that $k<<N$).

\subsection{Robust Enhancement via the Regularization Term}
\label{subsec:rob_enhcancement}

Our proposed scheme alternates between a regularized Least Squares step and an OMP selection step based on the residual. At this point, it should be noted that raw residuals may fail to detect outliers at leverage points; this is also known as swamping and masking of the outliers, \cite{huber1981wiley}. It is well known that this issue is strongly related to the input data. In our particular non-linear regularized setting, this occurs when the diagonal elements of the matrix $\bm{H}=\bm{X}_{(0)}(\bm{X}_{(0)}^T\bm{X}_{(0)})^{-1}\bm{X}_{(0)}^T$ obtain values close to one. Hence, many authors, e.g., \cite{huber1981wiley}, assume that $\max_{1\leq i \leq N} h_{ii} = h<<1$; however, in our setting, no such assumption is required since this can be accomplished via the use of the regularization term.
In the initialization step of KGARD, $\bm{X}_{(0)}=[\bm{K}\ \bm{1}]$ and $\bm{B}_{(0)}=\bm{I}_{N+1}$. Adopting the \emph{Singular Value Decomposition}\index{Singular Value Decomposition (SVD)} (SVD) for matrix $\bm{X}_{(0)}$, we obtain $\bm{X}_{(0)}=\bm{QSV}^T$, where $\bm{Q},\bm{V}$ are orthogonal, while $\bm{S}$ is the matrix of dimension $N \times (N+1)$ of the form $\bm{S}=\begin{bmatrix} \bm{\Sigma} & \bm{0} \end{bmatrix}$. The matrix $\bm{X}_{(0)}^T\bm{X}_{(0)}$ is positive semi-definite\index{Positive semi-definite}, thus all of its eigenvalues are non-negative. Hence, $\bm{\Sigma}$ is the diagonal matrix with entries the singular values\index{Singular values} of matrix $\bm{X}_{(0)}$, i.e.,  $\sigma_i \geq 0,\ i=1,...,N$.

At every iteration step, a regularized LS task is solved and
\begin{equation}
\bm{X}_{(0)}^T\bm{X}_{(0)}+\lambda \bm{I}_{N+1}=\bm{V} \underbrace{\begin{bmatrix}
\bm{\Sigma}^2 + \lambda \bm{I}_N & \bm{0} \\ \bm{0}^T & \lambda
\end{bmatrix}}_{\bm{\Lambda}} \bm{V}^T=\bm{V \Lambda V}^T.
\label{eq:Lambda_matrix}
\end{equation}
Hence, the new hat matrix is expressed as
\begin{equation}
\label{eq:hat_tilde_matrix}
\widetilde{\bm{H}}=\bm{X}_{(0)}(\bm{X}_{(0)}^T\bm{X}_{(0)}+\lambda \bm{I}_{N+1})^{-1}\bm{X}_{(0)}^T= \bm{QGQ}^T,
\end{equation}
where $\bm{G}$ is a diagonal matrix with
\begin{equation}
\bm{G}=\bm{\Sigma} (\bm{\Sigma}^2+\lambda \bm{I}_N)^{-1} \bm{\Sigma},\ g_{ii}=\frac{\sigma_i^2}{\sigma_i^2+\lambda},\ i=1,2,...,n,
\label{eq:Gmatrix}
\end{equation}
where $\lambda>0$ is the regularization parameter and $\sigma_i$ the  $i$-th singular value of the matrix $\bm{X}_{(0)}$. Hence, from \eqref{eq:hat_tilde_matrix}, it is a matter of simple manipulations to establish for the diagonal elements of the new hat matrix that they satisfy
\begin{equation}
\label{eq:regul_hat_matrix}
\tilde{h}_{ii} =\frac{\sigma_i^2}{\sigma_i^2+\lambda} h_{ii}.
\end{equation}
In simple words, the performed regularization down-weights the diagonal elements of the hat matrix. Equation \eqref{eq:regul_hat_matrix} is of major importance, since it guarantees that $\tilde{h}_{ii}< h_{ii}$ for any $\lambda > 0$. Furthermore, it is readily seen that as $\lambda \rightarrow 0$ the detection of outlier via the residual is forbidden, while as $\lambda \rightarrow \infty$ then $\tilde{h}_{ii} \rightarrow 0$ and thus occurrences of leverage points tend to disappear. In simple words, the regularization performed on the specific task guards the method against occurrences of leverage points. Of course, this fact alone does not guarantee that one could safely detect an outlier via the residual. This is due to the following two reasons: a) the values of the outliers could be too small (engaging with the inlier noise) or b) the fraction of outliers contaminating the data could be enormously large. Based on the previous discussion, we adopt the assumptions that the outliers are relatively few (the vector $\orub$ is sparse) and also that the outlier values are (relatively) large. From a practical point of view, the latter assumption is natural, since we want to detect values that greatly deviate from ``healthy" measurements. The first assumption is, also, in line with the use of the greedy approach. It is well established by now that greedy techniques work well for relatively small sparsity levels. These assumptions are also verified by the obtained experimental results.

\subsection{Further Improvements on KGARD's Performance}
\label{subsec:KGARD_improvemnts}
In order to simplify the theoretical analysis and reduce the corresponding equations, the proposed algorithm employs the same regularization parameter for all kernel coefficients. However, one may employ a more general scheme as follows:
\[
\begin{matrix}
\displaystyle{\min_{\bm{u},\bm{a}\in\mathbb{R}^N, c\in\mathbb{R}}}  & \|\bm{u}\|_0\\
\textrm{s. t.}  &  \|\bm{y} - \bm{K}\bm{a} - c\bm{1} - \bm{u}\|^2_2 + \|\bm{\Psi} \bm{a}\|_2^2 + \lambda c^2 \leq \varepsilon,
\end{matrix}
\]
where $\bm{\Psi}$ is a more general regularization matrix (Tikhonov matrix). For example, as the accuracy of kernel based methods usually drops near the border of the input domain, it is reasonable to increase the regularization effect at these points. This can be easily implemented by employing a diagonal matrix with positive elements on the diagonal (that correspond to $\lambda$) and increase the regularization terms that correspond to the points near the border. This is demonstrated in the experimental section \ref{sec:exper}.

\section{Theoretical Analysis - Identification of the Outliers}
\label{sec:theoretical_anal}

In the current section, the theoretical properties of the proposed robust kernel regression method, i.e., KGARD, are analyzed. In particular, we provide the necessary conditions so that the proposed method succeeds in identifying, first, the locations of all the outliers; the analysis is carried out for the case where only outliers are present in the noise. The derived theoretical condition for the second part (i.e., the outlier identification) is rather tight. However, as demonstrated in the experiments, the method achieves to recover the correct support of the sparse outlier vector in many cases where the theoretical result doesn't hold. This leads to the conclusion that the provided conditions can be loosen up significantly in the future. Moreover, in practice, where inlier noise also exists, the method succeeds to correctly identify the majority of the outliers. The reason that, the analysis is carried out for the case where inlier noise is not present, is due to the fact that the analysis gets highly involved. The absence of the inlier noise makes the analysis easier and it highlights some theoretical aspects on why the method works. It must be emphasized that, such a theoretical analysis is carried out for the first time and it is absent in the previously published works.


First of all, note that, for all $\varepsilon \geq 0$, there exists $\bm{z}$ such that $J_k({\bm{z}})\leq \varepsilon$. This implies that the feasible set of \eqref{eq:model_opt} is always nonempty\footnote{For example, if we select $\bm{z}=\left(\bm{0}^T,0,\bm{y}^T\right)^T$, then $J_k(\bm{z})=0$.}.
It is straightforward to prove that the set of \textit{normal equations}, obtained from \eqref{eq:J_opt}, at step $k$, is
\begin{equation}
(\bm{X}_{(k)}^{T}\bm{X}_{(k)} + \lambda \bm{B}_{(k)}) \bm{z}= \bm{X}_{(k)}^{T}\bm{y},
\label{eq:normal_equations}
\end{equation}
where $(\bm{X}_{(k)}^{T}\bm{X}_{(k)} + \lambda \bm{B}_{(k)})$ is invertible, i.e., \eqref{eq:J_opt} has a unique minimum, for all $k$. Recall that the matrix on the left side in \eqref{eq:normal_equations} is (strictly) positive definite, hence invertible. Alternatively, one could express \eqref{eq:J_opt} as follows:
\begin{align}
\begin{matrix}
\min_{\bm{z}}  & J_k(\bm{z}) = \left\|\left(\begin{matrix} \bm{y} \\ \bm{0}\end{matrix}\right) - \bm{D}_{(k)}\bm{z}\right\|_2^2,
\end{matrix}\label{eq:J_opt2}
\end{align}
where $\bm{D}_{(k)} = \begin{bmatrix}\bm{X}_{(k)}\\ \sqrt{\lambda}\bm{B}_{(k)} \end{bmatrix}$. Problem \eqref{eq:J_opt2} has a unique solution, if and only if the nullspaces of $\bm{X}_{(k)}$ and $\bm{B}_{(k)}$ intersect only trivially, i.e., $\mathcal{N}(\bm{X}_{(k)})\cap \mathcal{N}(\bm{B}_{(k)})=\{ 0 \}$ \cite{gander1978linear, gander1980least}. Hence, $\bm{M}_{(k)}=\bm{D}_{(k)}^T\bm{D}_{(k)}$ is (strictly) positive definite, as the columns of $\bm{D}_{(k)}$ are linearly independent and the minimizer $\bm{z}_{*} \in \Real^{N+1+k}$ of \eqref{eq:J_opt} is unique, \cite{bjorck1996numerical}. It should be emphasized that the matrix $\bm{B}$ is a projection matrix only for the regularization performed with the $\ell_2$-norm, thus equivalence between \eqref{eq:J_opt2} and \eqref{eq:J_opt} does not hold for the choice of matrix $\bm{B}$ with the kernel Gram matrix.

The following theorem establishes a bound on the largest singular value of matrix $\bm{X}_{(0)}$, which guarantees that the method first identifies the correct locations of all the outliers, for the case where only outliers exist in the noise. However, since the $\epsilon$ parameter controls the number of iterations, for which the method identifies an outlier, it is not guaranteed that it will stop once all the outliers are identified, unless the correct value is somehow given. Thus, it is possible that a few other locations, that do not correspond to outliers, are also identified. It must be pointed out that, such a result has never been established before by other comparative methods.

\begin{theorem}
Let $\bm{K}$ be a full rank, square, real valued matrix. Suppose, that
\[
\bm{y}=[\bm{K} \ \bm{1}] \underbrace{
(\ubar{\bm{\alpha}^T},\\ \ubar{c})^T}_{\ubar{\bm{\theta}}}  + \ubar{\bm{u}},
\]
where  $\ubar{\bm{u}}$ is a sparse (outlier) vector. KGARD is guaranteed to identify first the correct locations of all the outliers, if the maximum singular value of matrix $\bm{X}_{(0)}:= [\bm{K}\ \bm{1}]$, satisfies:
\begin{equation}
\label{eq:sigular_value_bound}
\sigma_{M}(\bm{X}_{(0)})< \gamma \sqrt{\lambda} ,
\end{equation}
\begin{flalign}
\text{where}\  \gamma &= \sqrt{ \frac{\min{|\ubar{u}|}- \sqrt{2\lambda} || \ubar{\bm{\theta}}||_2}{2||\ubar{\bm{u}}||_2 - \min{|\ubar{u}|}+ \sqrt{2\lambda} || \ubar{\bm{\theta}}||_2}},&
\label{eq:gamma}
\end{flalign}
$\min{|\ubar{u}|}$ is the smallest absolute value of the sparse vector over the non-zero coordinates and $\lambda>0$ is a sufficiently large\footnote{Since the regularization parameter is defined by the user, we assume that such a value can be achieved, so that the $\gamma$ parameter makes sense. More details can be found in the proof at the appendix section.} regularization parameter for KGARD.
\label{theor:sing_val_bound}
\end{theorem}
The proof is presented in the Appendix section.

\begin{remark}
Note that, the theorem does not guarantee that only the locations of the true outliers will be identified. If the value of $\epsilon$ is too small, then KGARD, once it  identifies the location of the true outliers, it will next identify locations that do not correspond to outlier indices.
\end{remark}

\section{Experiments}
\label{sec:exper}

For the entire section of experiments, the Gaussian (RBF) kernel is employed and all the results are averaged over 1000 ``Monte Carlo" runs (independent simulations). At each experiment, the parameters are optimized via cross-validation for each method; that is, exhaustive search is performed and the best values are chosen. Furthermore, the respective parameter values are given (for each method), so that results are reproducible. The MATLAB code can be found in \url{http://bouboulis.mysch.gr/kernels.html}.

\subsection{Identification of the Outliers}
\label{subsec:support}

\begin{figure}[t]
\centering
\includegraphics[scale=0.5]{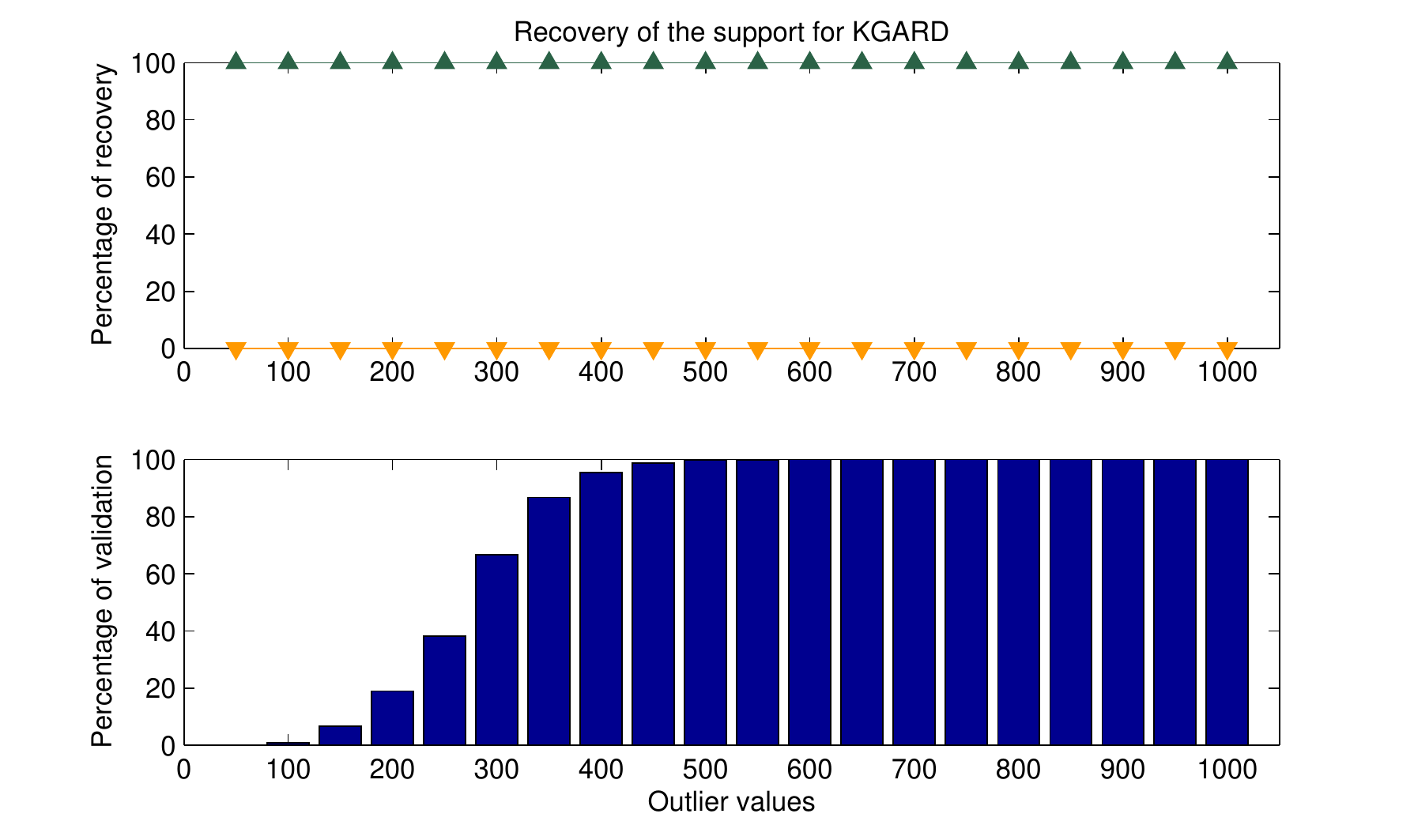}
\caption{Percentage of the correct (green pointing up) and wrong (orange pointing down) indices that KGARD has classified as outliers, while varying the values $\pm \oru$ of the outliers at the fixed fraction of 10\%. Although the condition \eqref{eq:sigular_value_bound} is valid only for values greater than $\pm 600$ (and with high probability valid for values 400-599), the support of the sparse outlier vector has been correctly estimated for smaller values of outlier noise as well.}
\label{FIG:supp_recovery}
\end{figure}

In the current section, our main concern is to test on the validity of the condition \eqref{eq:sigular_value_bound} in practise. To this end, we have performed the following experiment, for the case where only outliers exist in the noise.

%

We consider $N=100$ equidistant points over the interval $[0,1]$ and generate the output data via $\orf(x_i)=\sum_{j=1}^{N}\ora_j \kappa(x_i,x_j)$, where $\kappa$ is the Gaussian kernel with $\sigma=0.1$ and the vector of coefficients $\orab=[\ora_1,\dots,\ora_N]$ is a sparse vector with the number of non-zero coordinates ranging between 2 and 23 and their values drawn from $\mathcal{N}(0,{0.5}^2).$ Since no inlier noise exists, our corrupted data is given from \eqref{eq:model_ori} for $\eta_i=0$ and outlier values $\pm \oru$. Moreover, since the condition \eqref{eq:sigular_value_bound} is valid for fixed values of the parameters involved, we have measured KGARD's ability to successfully recover the sparse outlier vector's support, which is denoted by  $\mathcal{T}=\supp(\orub)$, while varying the values of the outliers. In Figure \ref{FIG:supp_recovery}, the identification of the outliers is demonstrated for KGARD. On the vertical axis, we have measured the percentage of correct and wrong (extra) indices that are classified as outliers, while varying their values $u$. In parallel, the bar chart demonstrates the validity of the introduced condition \eqref{eq:sigular_value_bound}. It is clear that, if the condition holds, KGARD identifies the correct support of the sparse outlier vector successfully. However, even if the condition is rarely satisfied, e.g., for smaller values such as $\oru=100$, the method still manages to identify the correct support. This fact leads to the conclusion that the condition imposed by \eqref{eq:sigular_value_bound} is rather strict. This is in line with most sparse modeling related conditions, which, in practice, fall short in predicting the exact recovery conditions.


It should also be noted that, experiments have been performed with the use of other types of non-linear functions, e.g. sinc, sinusoids, etc., and the results were similar to the ones presented here.

\begin{table}
\scriptsize
\renewcommand{\arraystretch}{0.25}
\centering
{\begin{tabular}{|c||c|c|c|c|} \thickhline
Method   & $MSE$ & Cor. - Wr. &  MIT & Noise \\ \thickhline

\textbf{RB-RVM}   & 0.0851 &  -   & 0.298 & 20 dB - 5\% \\\hline
$\begin{matrix} \textbf{RAM} \\ \lambda=0.07, \mu=2.5 \end{matrix}$ &   0.0345 &  100\% -  0.2\%  & 0.005 & 20 dB - 5\% \\\hline
$\begin{matrix} \textbf{KGARD} \\ \lambda=0.2, \varepsilon=10 \end{matrix}$ &  $\bm{0.0285}$ &  100\% - 0\% & 0.004 & 20 dB - 5\%  \\\hline

\textbf{RB-RVM}     & 0.0912 &  -   & 0.298 & 20 dB - 10\% \\\hline
$\begin{matrix} \textbf{RAM} \\ \lambda=0.07, \mu=2.5 \end{matrix}$ &   0.0372 &  100\% - 0.1\%  & 0.007 & 20 dB - 10\% \\\hline
$\begin{matrix} \textbf{KGARD} \\ \lambda=0.2, \varepsilon=10 \end{matrix}$ &  $\bm{0.0305}$ &  100 \% - 0 \% & 0.008 & 20 dB - 10\%  \\ \hline

\textbf{RB-RVM}  & 0.0994 &  -   & 0.299 & 20 dB - 15\% \\\hline
$\begin{matrix} \textbf{RAM} \\ \lambda=0.07, \mu=2 \end{matrix}$ &  0.0393 &  100\%  - 0.6\%  & 0.008 & 20 dB - 15\% \\\hline
$\begin{matrix} \textbf{KGARD} \\ \lambda=0.3, \varepsilon=10 \end{matrix}$ & $\bm{0.0330}$ &  100\%-  0\% & 0.012 & 20 dB - 15\%  \\ \hline

\textbf{RB-RVM}   & 0.1184 &  -  & 0.305 & 20 dB - 20\% \\\hline
$\begin{matrix} \textbf{RAM} \\ \lambda=0.07, \mu=2 \end{matrix}$ & $\bm{0.0422}$ &  100\%  - 0.4\%  & 0.010 & 20 dB - 20\% \\\hline
$\begin{matrix} \textbf{KGARD} \\ \lambda=1, \varepsilon=10 \end{matrix}$ &  0.0626 &  100\% - 0\% & 0.017 & 20 dB - 20\%  \\ \hline

\textbf{RB-RVM}  & 0.3631 &  - & 0.327 &  15 dB - 5\% \\\hline
$\begin{matrix} \textbf{RAM} \\ \lambda=0.15, \mu=5 \end{matrix}$  & 0.1036  &  100\%-   0.7\% &  0.005 &  15 dB - 5\% \\\hline
$\begin{matrix} \textbf{KGARD} \\ \lambda=0.3, \varepsilon=15 \end{matrix}$ & $\bm{0.0862}$  &  100\% -  0.1\% & 0.005 &  15 dB - 5\%  \\\hline

\textbf{RB-RVM}  & 0.3830 &  -  & 0.319 &  15 dB - 10\% \\\hline
$\begin{matrix} \textbf{RAM} \\ \lambda=0.15, \mu=5 \end{matrix}$ & 0.1118  &  100\% -  0.4 \% &  0.006 &  15 dB - 10\% \\\hline
$\begin{matrix} \textbf{KGARD} \\ \lambda=0.3, \varepsilon=15 \end{matrix}$  & $\bm{0.0925}$  &  100\% -  0\% & 0.008 &  15 dB - 10\%  \\\hline

\textbf{RB-RVM}   & 0.4166 &  -  & 0.317 &  15 dB - 15\% \\\hline
$\begin{matrix} \textbf{RAM} \\ \lambda=0.15, \mu=5 \end{matrix}$ & 0.1186  &  100\%  - 0.3\% &  0.007 &  15 dB - 15\% \\\hline
$\begin{matrix} \textbf{KGARD} \\ \lambda=0.3, \varepsilon=15 \end{matrix}$  & $\bm{0.1003}$  &  100\% -  0\% & 0.012 &  15 dB - 15\%  \\\hline

\textbf{RB-RVM}  & 0.4798 &  - & 0.312 &  15 dB - 20\% \\\hline
$\begin{matrix} \textbf{RAM} \\ \lambda=0.15, \mu=4 \end{matrix}$ & $\bm{0.1282}$  &  100\%   - 1.4 \% &  0.008 &  15 dB - 20\% \\\hline
$\begin{matrix} \textbf{KGARD} \\ \lambda=0.7, \varepsilon=15 \end{matrix}$ & 0.1349  &  100\% -  0\% & 0.016 &  15 dB - 20\%  \\\hline
\end{tabular}}
\caption{MSE for $\orf(x)=20\sinc(2\pi x)$ computed over the validation set, percentage of correct and wrong support recovered and mean implementation time (MIT) in seconds, for each level of inlier noise and fraction of outliers.}
\label{TAB:MSE_1D_SINC}
\end{table}

\begin{table}
\scriptsize
\renewcommand{\arraystretch}{0.25}
\centering
{\begin{tabular}{|c||c|c|c|c|} \thickhline
Method  &    $MSE$ & Cor. -  Wr. supp &  MIT (sec) & Outliers \\ \thickhline

\textbf{RB-RVM} & 3.6918 &  -  & 0.416 &   5\% \\\hline
$\begin{matrix} \textbf{RAM} \\ \lambda=0.2, \mu=22 \end{matrix}$  & 1.8592  &  100\%  -  0.1 \% &  0.010 &  5\% \\\hline
$\begin{matrix} \textbf{KGARD} \\ \lambda=0.15, \varepsilon=46 \end{matrix}$  & $\bm{1.5644}$  &  100 \% -  0.3 \% & 0.009 &   5\%  \\\hline

\textbf{RB-RVM}   & 3.8977 &  -  & 0.419 &  10\% \\\hline
$\begin{matrix} \textbf{RAM} \\ \lambda=0.2, \mu=18 \end{matrix}$ & 1.9926  &  100\%  -  0.9 \% &  0.013 &  10\% \\\hline
$\begin{matrix} \textbf{KGARD} \\ \lambda=0.15, \varepsilon=44 \end{matrix}$  & $\bm{1.6750}$  &  100 \%  -  0.5 \% & 0.016 &  10\%  \\\hline

\textbf{RB-RVM}   & 4.2181 &  -  & 0.418 &  15\% \\\hline
$\begin{matrix} \textbf{RAM} \\ \lambda=0.2, \mu=17 \end{matrix}$ & 2.2846  &  100\%  -  1.6 \% &  0.016 &  15\% \\\hline
$\begin{matrix} \textbf{KGARD} \\ \lambda=0.2, \varepsilon=42 \end{matrix}$ & $\bm{1.9375}$  &  99.9 \%  -  0.9 \% & 0.024 &  15\%  \\\hline
\textbf{RB-RVM}   & 5.0540 &  -  & 0.418 &  20\% \\\hline
$\begin{matrix} \textbf{RAM} \\ \lambda=0.2, \mu=16 \end{matrix}$ & 2.6703  &  99.9\%  -  2.3 \% &  0.020 &  20\% \\\hline
$\begin{matrix} \textbf{KGARD} \\ \lambda=0.4, \varepsilon=42 \end{matrix}$ & $\bm{2.6113}$ &  99.9 \%  -  1 \% & 0.033 &  20\%  \\\hline

\end{tabular}}
\caption{Performance evaluation for each method, for the case where the input data lies on the two-dimensional space and the output $\orf \in \mathcal{H}$ is considered as a linear combination of a few kernels. The inlier noise is considered random Gaussian with $\sigma=3$. For each fraction of outliers, the MSE over the validation set, the percentage of correct and wrong locations that the method has identified and the mean implementation time (MIT), are listed.}
\label{TAB:MSE_2D_Linear Kernel}
\end{table}

\subsection{Evaluation of the Method: Mean-Square-Error (MSE)}
\label{subsec:mse_comparison}

In the current section, the previously established methods that deal with the non-linear robust estimation with kernels, i.e., the Bayesian approach RB-RVM and the weighted $\ell_1$-norm approximation method (RAM), are compared against KGARD in terms of the mean-square-error (MSE) performance. Additionally, the evaluation is enhanced with a list of the percentage of the correct and wrong indices that each method has classified as outliers, for all methods except for the Bayesian approach (not directly provided by the RB-RVM method). Moreover, the \textit{mean implementation time} (MIT) is measured for each experiment.
Finally, according to Section \ref{subsec:KGARD_improvemnts}, for the first experiment (one-dimensional case), we have increased the regularization value $\lambda$ of KGARD near the edge points/borders, as a means to improve its performance.
In particular, at the 5 first and 5 last points (borders), the regularizer is
automatically multiplied by the factor of 5, with respect to the predefined
value $\lambda$ which is used on the interior points.
The experiments are described in more detail next.

For the first experiment, we have selected the $\sinc$ function, due to its well established properties in the context of signal processing. We have considered $398$ equidistant points over the interval $[-0.99,1)$ for the input values and generated the uncorrupted output values via $\orf(x_i)=20\sinc(2 \pi x_i)$. Next, the set of points is split into two subsets, the training and the validation subset. The training subset, with points denoted by $(y_i, x_i)$, consists of the $N=199$ odd indexed points (first, third, etc.), while the validation subset comprises the remaining points (denoted as $(y'_i,x'_i)$). The original data of the training set, is then contaminated by noise, as \eqref{eq:model_ori} suggests. The inlier part is considered to be random Gaussian noise of appropriate variance (measured in dB), while the outlier part consists of various fractions of outliers, with constant values $\pm 15$, distributed uniformly over the support set. Finally, the kernel parameter $\sigma$ has been set equal to $\sigma=0.15$.
Table \ref{TAB:MSE_1D_SINC} depicts each method's performance, where the best results are marked in \textbf{bold}. In terms of the computed MSE over the validation/testing subset, it is clear that KGARD attains the lowest validation error for all fractions of outliers, except for the fraction of 20\%.
This fact is also in line with the theoretical properties of the sparse greedy methods, since their performance boosts as the sparsity level of the approximation is relatively low. On the other hand, the RAM solver seems more suitable for larger fractions of outliers. Moreover, the computational cost is comparable for both methods (RAM and KGARD), for small fractions of outliers. Regarding the identification of the sparse outlier vector support, although both methods correctly identify the indices that belong to the sparse outlier vector's support, i.e., $\mathcal{T}=\supp(\orub)$, RAM (incorrectly) classifies more indices as outliers than KGARD.

\begin{table}
\small
\renewcommand{\arraystretch}{0.25}
\centering
{\begin{tabular}{|c||c|c|c|} \thickhline
Method  &    $MSE$  &  MIT (sec) & Noise Parameters \\ \thickhline

\textbf{RB-RVM} & 0.0435  & 0.305 & $\alpha =1.2,\ \gamma = 0.3$   \\\hline
$\begin{matrix} \textbf{RAM} \\ \lambda=0.04, \mu=0.4 \end{matrix}$ &  0.0270 &  0.017 &  $\alpha =1.2,\ \gamma = 0.3$   \\\hline 
$\begin{matrix} \textbf{KGARD} \\ \lambda=0.18, \varepsilon=5 \end{matrix}$   &  $\bm{0.0261}$  & 0.011 &   $\alpha =1.2,\ \gamma = 0.3$    \\\hline

\textbf{RB-RVM} & 1.7037  & 0.295 & $\alpha =1.2,\ \gamma = 2$   \\\hline
$\begin{matrix} \textbf{RAM} \\ \lambda=0.19, \mu=5 \end{matrix}$ &  $\bm{0.8978}$ &  0.010 &  $\alpha =1.2,\ \gamma = 2$   \\\hline 
$\begin{matrix} \textbf{KGARD} \\ \lambda=1.2, \varepsilon=32 \end{matrix}$   &  0.9171  & 0.011 &   $\alpha =1.2,\ \gamma = 2$    \\\hline

\end{tabular}}

\caption{Performance evaluation for each method, for the case where the input data lies on the two-dimensional space and the output $\orf \in \mathcal{H}$ is considered as a linear combination of a few kernels. The inlier noise is considered random Gaussian with $\sigma=3$. For each fraction of outliers, the MSE over the validation set, the percentage of correct and wrong locations that the method has identified and the mean implementation time (MIT), are listed.}
\label{TAB:MSE_1D_Heavy_tailed}
\end{table}

For the second experiment, KGARD's performance is tested for the case where the input data lies on a two-dimensional subspace. To this end, we consider $31$ points in $[0,1]$ and separate these points, to form the training set, which comprises 16 odd indices and the rest 15, forming the validation set. Next, the $31^2$ points are distributed over a squared lattice in plane $[0,1] \times [0,1]$, where each uncorrupted measurement is  generated by $\orf(\bm{x}_i) =\sum_{j=1}^{31^2}\ora_j \kappa (\bm{x}_i, \bm{x}_j)$, ($\sigma=0.2$) and a sparse coefficient vector $\orab=[\ora_1,\dots,\ora_{31}]$ with non-zero values ranging between $4\% - 17.5 \%$ and their values randomly drawn from $\mathcal{N}(0,{25.6}^2).$ Thus, the training subset, consists of $N=16^2$ points, while the remaining $15^2$ correspond to the validation/testing subset. According to equation \eqref{eq:model_ori}, the original observations of the training set are corrupted by inlier noise originating from $\mathcal{N}(0,3^2)$ and outlier values $\pm 40$. The results are given in Table \ref{TAB:MSE_2D_Linear Kernel} for various fractions of outliers, with the best values of the (validation) MSE marked in \textbf{bold}. It is evident that, for the two-dimensional non-linear denoising task, KGARD's performance outperforms its competitors (in terms of MSE), for all fractions of the outliers.

Finally, it should also be noted that, although RB-RVM does not perform as good as its competitors, it has the advantage that no parameter tuning is required; however, this comes at substantially increased computational cost. On the contrary, the pair of tuning parameters for RAM, renders the method very difficult to be fully optimized (in terms of MSE), in practise. In contrast, taking into account the physical interpretation of $\epsilon$ and $\lambda$ associated with KGARD, we have developed a method for automatic user-free choice of these variables in the image denoising task.

\subsection{Simulations with Noise Originating from a Heavy-tailed Distribution}

Finally, we have experimented with more general types of noise, i.e., noise originating from a heavy-tailed distribution. In particular, we have considered that the noise variable in \eqref{eq:nonl_equation_inlier_only} belongs to the L\'{e}vy alpha-stable distribution, with pdf expressed in closed form only for special cases of the involved parameters, $\alpha,\ \beta,\ \gamma$ and $\delta$. The distribution's parameter $\beta$ controls the \textit{skewness} and is set to zero (results to a symmetric distribution without skewness). The parameter $\alpha \in (0,2]$ is called the \textit{characteristic exponent} and describes the tail of the distribution, while $\gamma>0$ is the \textit{scale} parameter and $\delta \in \Real$ is the \textit{location}. The last two parameters are similar to what the variance and the mean are to the normal distribution. In our setup, we have considered the location centered around zero ($\delta=0$). The rest of the variables are listed in Table \ref{TAB:MSE_1D_Heavy_tailed} for each experiment that has been performed. There, we have listed the validation MSE (second column) as well as the optimum values of the involved parameters and the mean implementation time (third column). It is readily seen that for the first experiment $(\gamma=0.3)$ KGARD achieves the lowest MSE, in less time. However, in the second experiment ($\gamma=2$), RAM outperforms KGARD. This result is to no surprise, since greedy methods are not expected to excel in very noisy setups.

\section{Application in Image Denoising}\label{sec:image_den}
In this section, in order to test the proposed algorithmic scheme in real-life scenarios, we use the KGARD framework to address one of the most popular problems that rise in the field of image processing: the task of removing noise from a digital image. The source of noise in this case can be either errors of the imaging system itself, errors that occur due to limitations of the imaging system, or errors that are generated by the environment. Typically, the noisy image is modeled as follows:
$g(x,x') = \org(x,x') + v(x,x'),$
for $x, x' \in [0, 1]$, where $\org$ is the original noise-free image and $v$ the additive noise. Given the noisy image $g$, the objective of any image denoising method is to obtain an estimate of the original image $\org$.  In most cases, we assume that the image noise is Gaussian additive, independent at each pixel, and independent of the signal intensity, or that it contains spikes or impulses (i.e., salt and pepper noise). However, there are cases where the noise model follows other probability density functions (e.g., the Poisson distribution or the uniform distribution).
Although the wavelet-based image denoising methods have dominated the research (see for example \cite{Portilla2003, BiShrink2, DaFoKatEg}), there are other methods that can be employed successfully, e.g., methods based on Partial Differential Equations, neighborhood filters, the non local means rationale, read \cite{milanfar2013tour, talebi2014global}, and/or non linear modeling using local expansion approximation techniques, \cite{TaFaMil}. The majority of the aforementioned methods assume a specific type of noise model. In fact, most of them require some sort of a priori knowledge of the noise distribution.  In contrast to this approach, the more recently introduced denoising methods based on KRR make no assumptions about the underlying noise model and, thus, they can effectively treat more complex models, \cite{bouboulis2010adaptive}.

In this section, we demonstrate how the proposed KGARD algorithmic scheme can be used to treat the image denoising problem in cases where the noise model includes impulses. We will present two different denoising methods to deal with this type of noise. The first one is directly based on KGARD algorithmic scheme, while the second method splits the denoising procedure into two parts: the identification and removal of the impulses is first carried out, via the KGARD and then the output is fed into a cutting edge wavelet based denoising method to cope with the bounded noise component.


\subsection{Modeling the Image and the Noise}\label{subsec:image_den:model}
\begin{figure}
\begin{tabular}{cc}
\includegraphics[scale=0.22]{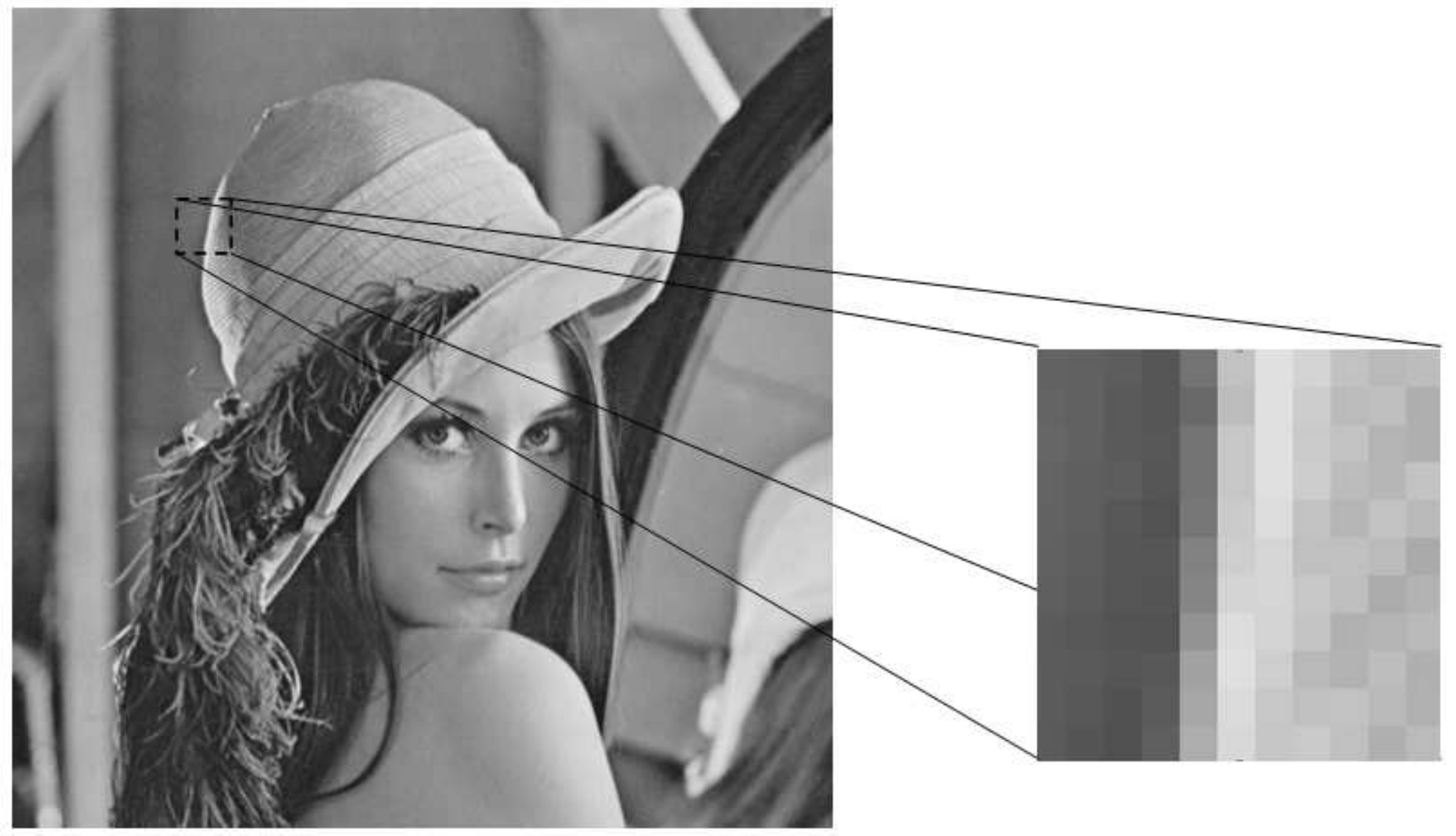} & \includegraphics[scale=0.22]{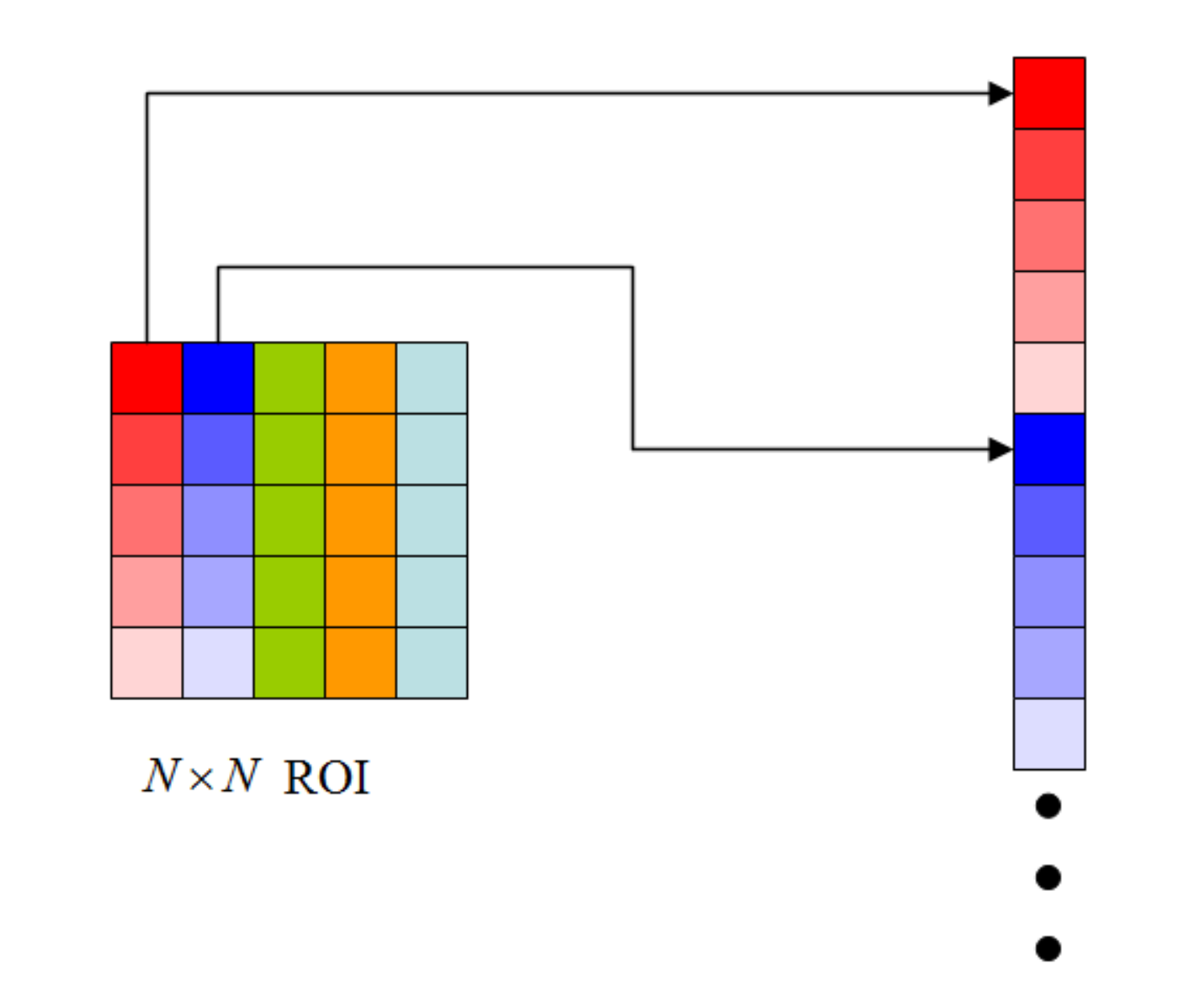}\\
(a)  &   (b)
\end{tabular}
\caption{(a) A square $N \times N$ region of intest (ROI).  (b) Rearranging the pixels of a ROI.}\label{FIG:im_region}
\end{figure}

In the proposed denoising method, we adopt the well known and popular strategy of dividing the ``noisy'' image into smaller $N\times N$ square regions of interest (ROIs), as it is illustrated in
Figure \ref{FIG:im_region}. Then, we rearrange the pixels so that to form a row vector. Instead of applying the denoising process to the entire image, we process each ROI individually in sequential order. This is done for two reasons: (a) Firstly, the time needed to solve the optimization tasks considered in the next sections increases polynomially with $N^2$ and (b) working with each ROI separately enables us to change the parameters of the model in an adaptive manner, to account for the different level of details in each ROI. The rearrangement shown in Figure \ref{FIG:im_region} implies that, the pixel $(i,j)$ (i.e., $i$-th row, $j$-th column) is placed at the $n$-th position of the respective vector, where $n=(i-1)\cdot N + j$.

In KRR denoising methods, one assumes that each ROI represents the points on the surface of a continuous function, $\org$, of two variables defined on $[0, 1] \times [0, 1]$. The pixel values of the noise-free and the noisy digitized ROIs are represented as $\ubar{\zeta}_{ij} = \org(x_i, x'_j)$ and $\zeta_{ij}$ respectively (both taking values in the interval $[0, 255]$), where $x_i = (i-1)/(N-1)$, $x'_j = (j-1)/(N-1)$, for $i,j = 1, 2, ...,N$. Moreover, as the original image $\org$ is a relatively smooth function (with the exception close to the edges), we assume that it lies in an RKHS induced by the Gaussian kernel, i.e., $\org\in\cal{H}$, for some $\sigma>0$. Specifically, in order to be consistent with the representer theorem, we will assume that $\org$ takes the form of a finite linear representation of kernel functions centered at all pixels, thus after pixel rearrangement we can write:
\begin{align}
\org = \sum_{n=1}^{N^2} \ora_{n} \kappa(\cdot,\bm{x}_n), \label{eq:image_repr2}
\end{align}
where $\bm{x}_n = (x_i, x'_j)$ and $n=(i-1)\cdot N + j$.
Hence, the intensity of the $n$-th pixel is given by
\begin{align}
\ubar{\zeta}_n = \org(\bm{x}_n) = \sum_{m=1}^{N^2} \ora_{m} \kappa(\bm{x}_n,\bm{x}_m).\label{eq:pixel_int}
\end{align}

The model considered in this paper assumes that the intensity of the pixels of the noisy ROI can be decomposed as
$
\zeta_{ij} =  \ubar{\zeta}_{ij} + \oru_{ij} + \eta_{ij},$
for $i,j = 1, 2, ...,N$, where $\eta_{ij}$ denotes the bounded noise component and $\oru_{ij}$ the possible appearance of an outlier at that pixel. In vector notation (after rearrangement), we can write
$\bm{\zeta} = \ubar{\bm{\zeta}} + \orub + \bm{\eta},$
where $\ubar{\bm{\zeta}}, \bm{\zeta}, \orub,\bm{\eta},  \in\mathbb{R}^{N^2}$, $\|\bm{\eta}\|_2\leq \epsilon$ and $\orub$ is a sparse vector. Moreover, exploiting \eqref{eq:pixel_int}, we can write $\ubar{\bm{\zeta}} = \bm{K}\cdot \orab$, where $\kappa_{nm} = \kappa(\bm{x}_n, \bm{x}_m)$. In this context, we can model the denoising task as the following optimization problem:
\begin{align}
\small
\begin{matrix}
\displaystyle{\min_{\bm{a}, \bm{u} \in \mathbb{R}^{N^2}, c \in \Real}}  &   \|\bm{u}\|_0\\
\textrm{s. t.}  &   \|\bm{\zeta} - \bm{K}\bm{a}-c\bm{1} - \bm{u}\|^2_2 + \lambda\|\bm{a}\|^2_2 + \lambda c^2 \leq \varepsilon,
\end{matrix}\label{eq:denoi_task}
\end{align}
for some predefined $\lambda,\varepsilon>0$. In a nutshell, problem (\ref{eq:denoi_task}) solves for the sparsest outlier's vector $\bm{u}$ and the respective $\bm{a}$ (i.e., the coefficients of the kernel expansion) that keep the error low, while at the same time preserve the smoothness of the original noise-free ROI (this is done via the regularization of the constraint's inequality). The regularization parameter $\lambda$ controls the smoothness of the solution. The larger the $\lambda$ is, the smoother the solution becomes, i.e., $\hat{\bm{\zeta}} = \bm{K} \hat{\bm{\alpha}}$.


\subsection{Implementation}\label{sec:algo_denoi}
The main mechanism of both algorithms that are presented in this section is simple. The image is divided into $N\times N$ ROIs and the KGARD algorithm is applied in each individual ROI sequentially. However, as the reconstruction accuracy drops near the borders of the respective domain, we have chosen to discard the values at those points. This means that although KGARD is applied to the $N\times N$ ROI, only the $L\times L$ values are used in the final reconstruction (those that are at the center of the ROI). In the sequel, we will name the $L\times L$ centered region as ``reduced ROI'' or rROI for short.
We will also assume that the dimensions of the image are multipliers of $L$ (if they are not, we can add dummy pixels to the end) and select $N$ so that $N-L$ is an even number.

After the reconstruction of a specific rROI, the algorithm moves to the next one, i.e., it moves $L$ pixels to the right, or, if the algorithm has reached the right end of the image, it moves at the beginning of the line, which is placed $L$ pixels below. Observe that, for this procedure to be valid, the image has to be padded by adding $(N-L)/2$ pixels along all dimensions. In this paper, we chose to pad the image by repeating border elements\footnote{This can be done with the ``replicate" option of MatLab's function \textit{padarray}.}. For example, if we select $L=8$ and $N=12$ to apply this procedure on an image with dimensions\footnote{Observe that $L$ divides $32$.} $32\times 32$, we will end up with a total of $16$ overlapping ROIs, $4$ per line.


Another important aspect of the denoising algorithm is the automated selection of the parameters $\lambda$ and $\epsilon$, that are involved in KGARD. This is an important feature, as these parameters largely control both the quality of the estimation and the recovery of the outliers and have to be tuned for each specific ROI. Naturally, it would have been intractable to require a user pre-defined pair of values (i.e., $\lambda, \epsilon$) for each specific ROI. Hence, we devised simple heuristic methods to adjust these values in each ROI depending on its features.

\subsubsection{Automatic selection of the regularization parameter $\lambda$}\label{subsubsec:lambda_adjust}
This parameter controls the smoothing operation of the denoising process. The user enters a specific value for $\lambda_0$ to control the strength of the smoothening and then the algorithm adjusts this value at each ROI separately, so that
$\lambda$ is small at ROIs that contain a lot of ``edges'' and large at ROIs that contain smooth areas. Whether a ROI has edges or not is determined by the mean magnitude of the gradient at each pixel. The rationale is described below:

{\footnotesize
\begin{itemize}
\item Select a user-defined value $\lambda_0$.
\item Compute the magnitude of the gradient at each pixel.
\item Compute the mean gradient of each ROI, i.e., the mean value of the gradient's magnitude of all pixels that belong to the ROI.
\item Compute the mean value, $m$, and the standard deviation, $s$, of the aforementioned mean gradients.
\item ROIs with mean gradient larger than $m+s$ are assumed to be areas with fine details and the algorithm sets $\lambda=\lambda_0$.
\item All ROIs with mean gradient lower than $m-s/10$ are assumed to be smooth areas and the algorithm sets $\lambda=15 \lambda_0$.
\item For all other ROIs the algorithm sets $\lambda=5\lambda_0$.
\end{itemize}
}

\subsubsection{Automatic computation of the termination parameter $\epsilon$}\label{subsubsec:epsilon_adjust}
\begin{figure}
\centering
\begin{tabular}{cc}
\includegraphics[scale=0.25]{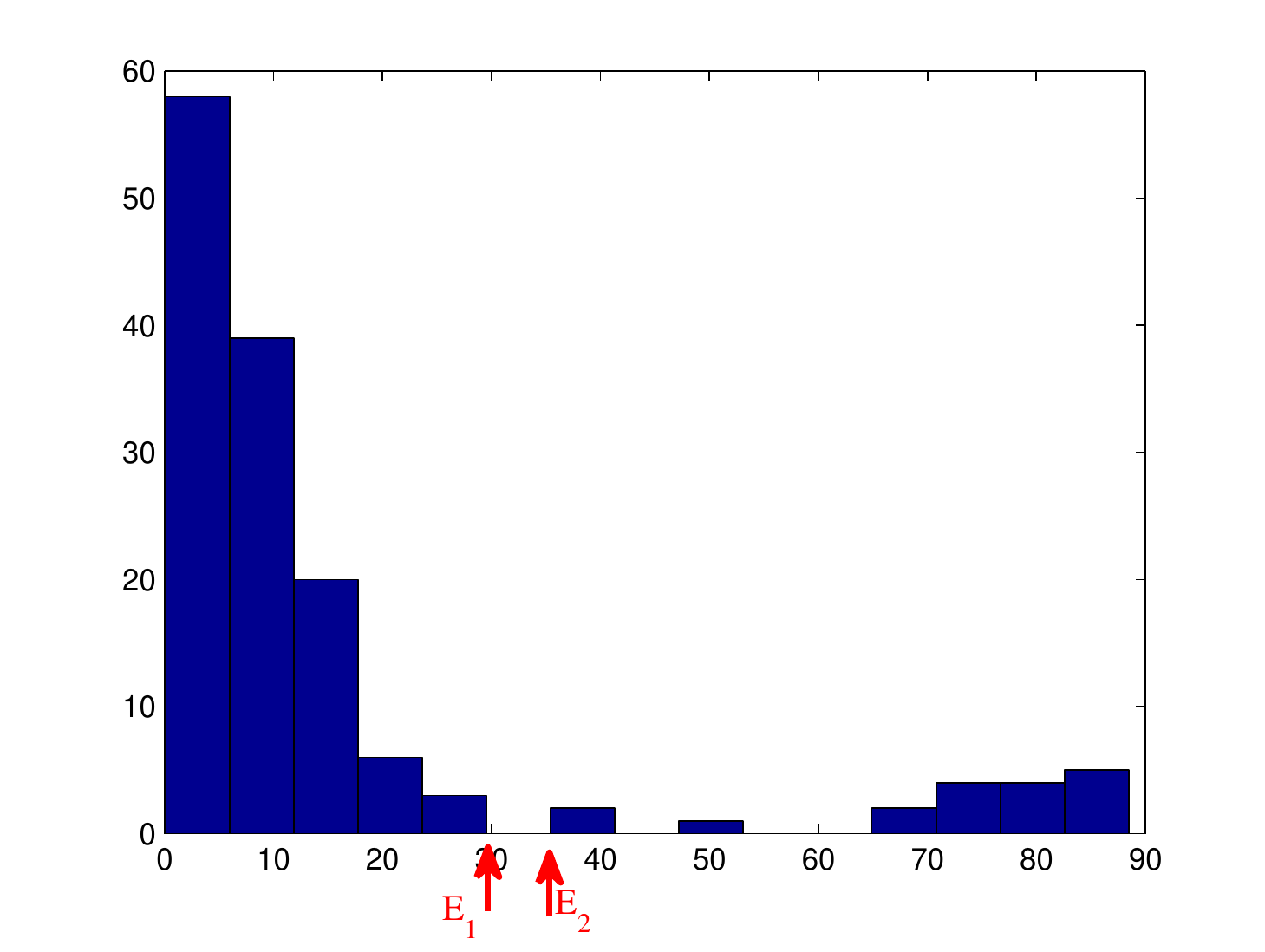} & \includegraphics[scale=0.25]{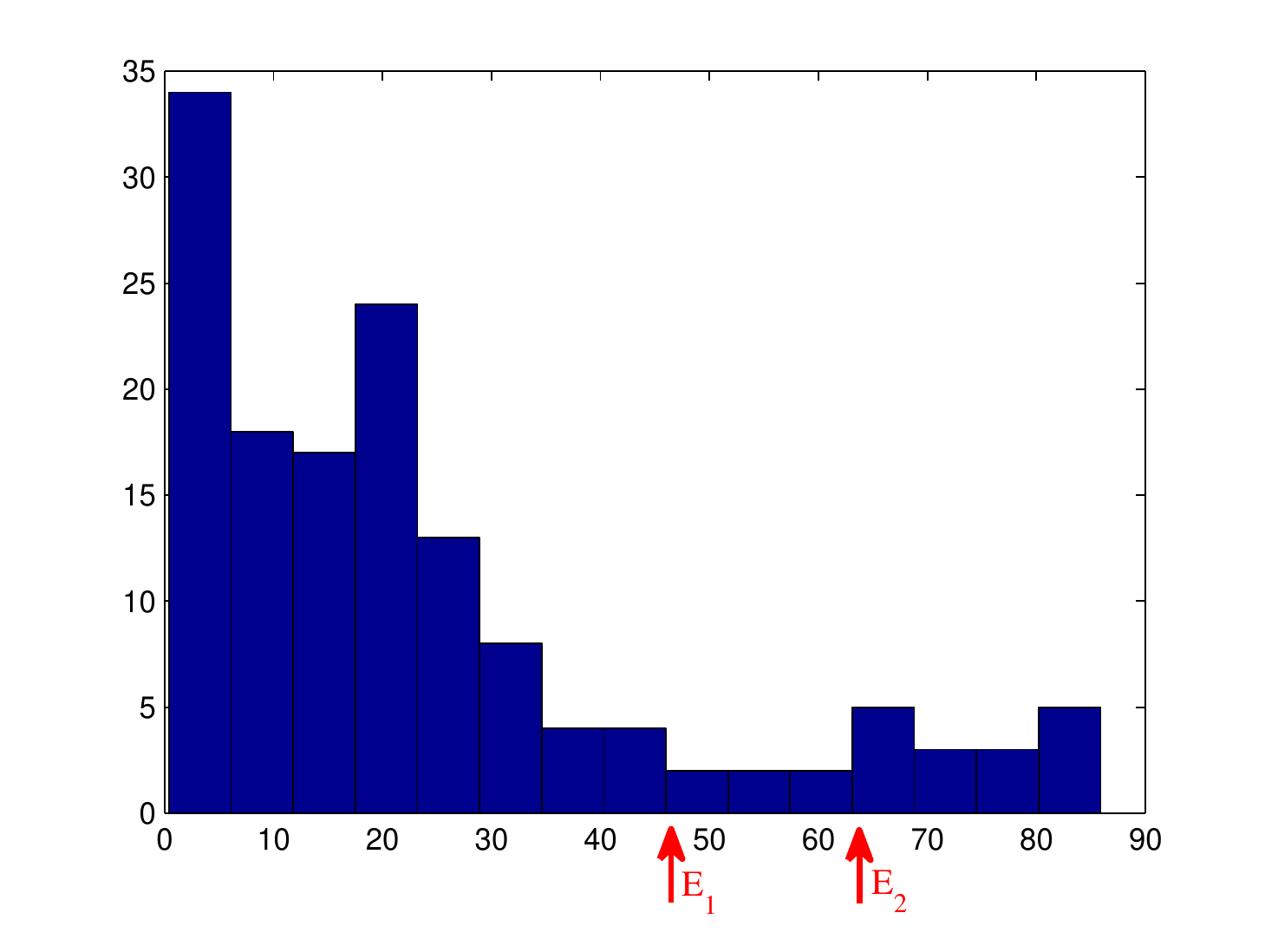}\\
(a) & (b)\\
\includegraphics[scale=0.25]{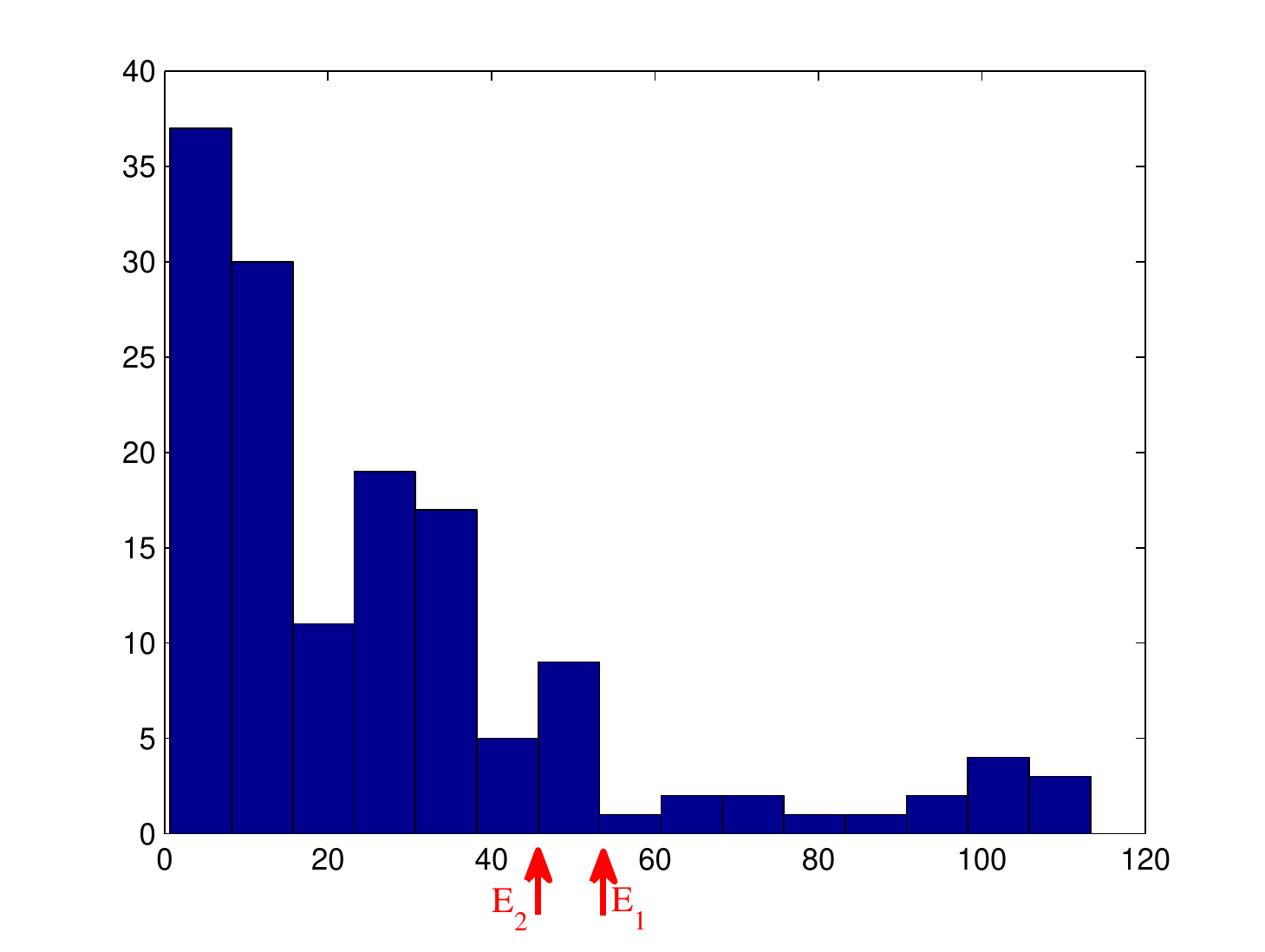} & \includegraphics[scale=0.25]{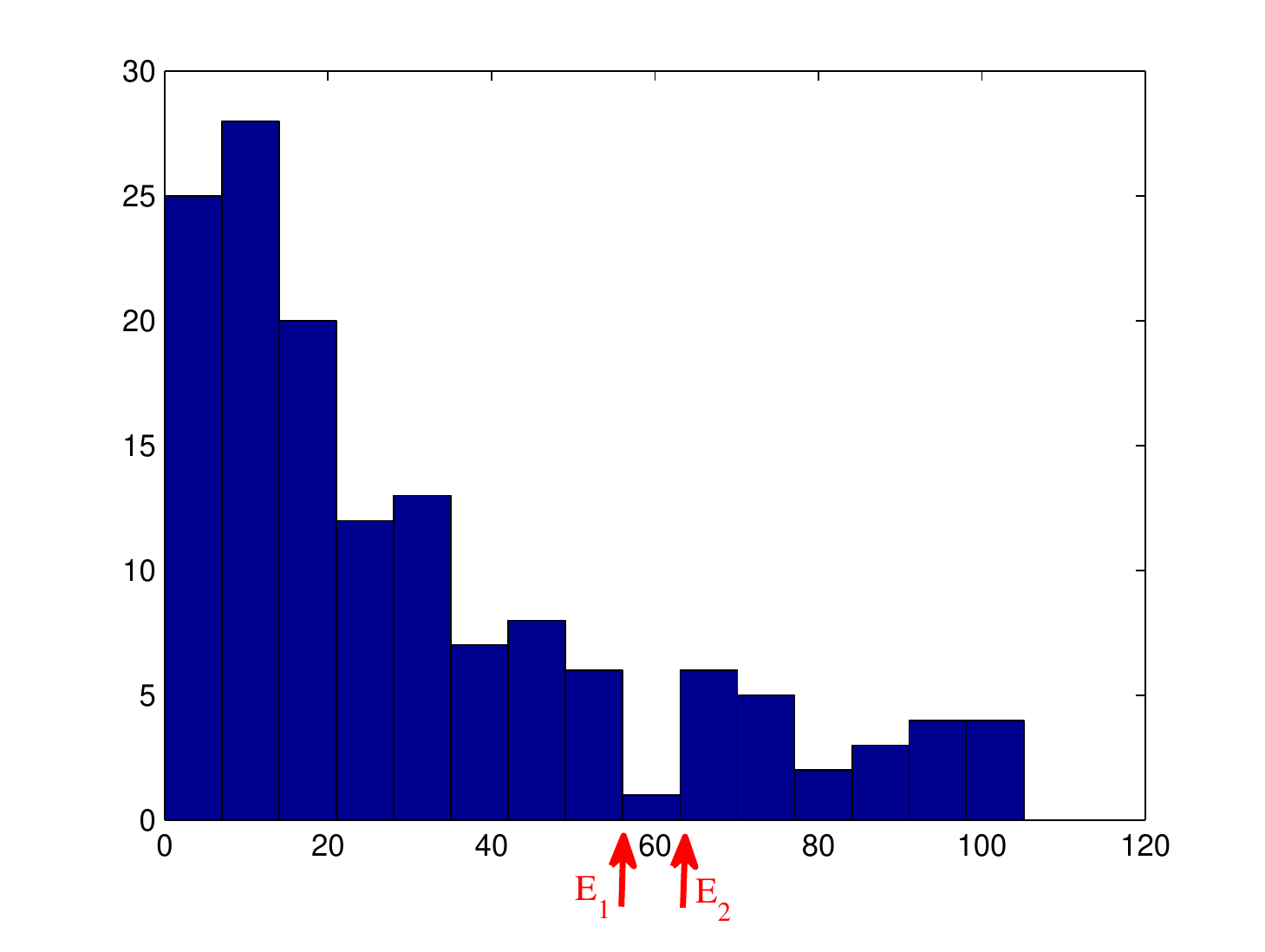}\\
(c) & (d)\\
\end{tabular}
\caption{Histogram of the residual's absolute value and the evaluation of $E_1$, $E_2$. (a), (b) and (d): $E_1$ is on the left of $E_2$. (c):  $E_1$ is on the right of $E_2$.}
\label{fig:histogram}
\end{figure}

In the image denoising case, the stopping criterion of KGARD is slightly modified. Hence, instead of requiring the norm of the residual vector to drop below $\epsilon$, i.e., $\|\bm{r}_{(k)}\|_2 \leq \epsilon$, we require the maximum absolute valued coordinate of $\bm{r}_{(k)}$ to drop below $\epsilon$ ($\left\|\bm{r}_{(k)} \right\|_{\infty}\leq \epsilon$). The estimation of $\epsilon$, for each particular ROI, is carried out as follows. Initially, a user-defined parameter $E_0$ is selected. At each step, a histogram chart with elements $|r_{(k),i}|$ is generated, using $\left[\frac{N^2}{10}\right]+1$ equally spaced bins along the $x$-axis, between the minimum and maximum values of $|r_{(k),i}|$. Note that, if outliers are present, then this histogram will have two distinct parts. One at the far right end representing the outliers and another (left) representing the pixels where the model fits well, as demonstrated in Figure \ref{fig:histogram}. This is the main principle that underlies the heuristic rules presented here. Let $\bm{h}$ denote the heights of the bars of the histogram and  $h_m$ be the minimum height of the histogram bars. Next, two real numbers, i.e., $E_1$, $E_2$, are defined. In particular, the number $E_1$ represents the left endpoint of the first occurrence of a minimum-height bar (i.e., the first bar with height equal to $h_m$, moving from left to right). The number $E_2$ represents the left endpoint of the first bar, $\ell$, with height $h_\ell$ (moving from left to right) that satisfies both $h_{\ell} - h_{\ell-1}\geq 1$ and $h_{\ell-1}\leq h_m + 5,\ \ell \geq 2 $. This roughly corresponds to the first increasing bar, which in parallel is next to a bar with height close to the minimum height. Both $E_1$ and $E_2$ are reasonable choices for the value of $\epsilon$ (meaning that the bars to the right of these values may be assumed to represent outliers). Finally, the algorithm determines whether the histogram can be clearly divided into two parts; the first one represents the usual errors and the other the errors due to outliers by using a simple rule: if $\frac{\sqrt{\textrm{var}(\bm{h}_{(k)})}}{\textrm{mean}(\bm{h}_{(k)})}>0.9$, then the two areas can be clearly distinguished, otherwise it is harder to separate these areas. Note that, we use the notation $\bm{h}_{(k)}$ to refer to the heights of the histogram bar at the $k$ step of the algorithm. The final computation of $\epsilon$ (at step $k$) is carried out as follows:
\begin{align}
\epsilon_{(k)} = \left\{ \begin{matrix} \min\{E_0, E_1, E_2\}, & \textrm{if } \frac{\sqrt{\textrm{var}(\bm{h}_{(k)})}}{\textrm{mean}(\bm{h}_{(k)})}>0.9 \\
\min\{E_0, E_1\}, & \textrm{otherwise}.\end{matrix}  \right.
\label{eq:e_auto_criterion}
\end{align}
It should be noted that, the user defined parameter $E_0$ has little importance in the evaluation of $\epsilon$. One may set it constantly to a value near $40$ (as we did in all provided simulations). However, in cases where the image is corrupted by outliers only, a smaller value may be advisable, although it does not have a great impact on the reconstruction quality.

\subsubsection{Direct KGARD implementation}

The first denoising method, which we call KGARD for short, is described in Algorithm \ref{alg:algo_d1}. The algorithm requires five user-defined parameters: (a) the regularization parameter, $\lambda_0$, (b) the Gaussian kernel width, $\sigma$, (c) the OMP termination parameter $\epsilon$, (d)  the size of the ROI, $N$ and (e) the size of the rROIs, that are used in the reconstruction, i.e., $L$. However, these parameters are somehow interrelated. We will discuss these issues in the next sections.

\begin{algorithm}[t]
\footnotesize
\caption{KGARD for image denoising}\label{alg:algo_d1}
\begin{algorithmic}[1]
\State \textbf{Input}: the original noisy image $\bm{I}$ and $\lambda_0$, $\sigma$, $E_0$, $N$, $L$.
\State \textbf{Output}: the denoised image $\hat{\bm{I}}$ and the outliers' image $\hat{\bm{O}}$.

\State Build the kernel matrix $\bm{K}$.
\If {the dimensions of the original image are not multiplies of $L$}
\State Add initial padding
\EndIf
\State Form  $\hat{\bm{I}}$ and $\hat{\bm{O}}$ to the same dimensions as $\bm{I}$.
\State Add padding with size $N-L$ around the image.
\State Divide the image into $N\times N$ ROIs and compute the regularization parameters for each ROI.
\For{each ROI $\bm{ R}$}
\State Rearrange the pixels of $\bm{ R}$ to form the vector $\bm{\zeta}$.
\State Run the modified KGARD algorithm on the set $\bm{\zeta}$ with parameter $\lambda$ and stoping criterion as described in section \ref{subsubsec:epsilon_adjust}.
\State Let $\hat{\bm{a}}$, $\hat{\bm{u}}$ be the solution of KGARD.
\State Compute the denoised vector $\hat{\bm{\zeta}} = \bm{K} \hat{\bm{a}}$.
\State Rearrange the elements of  $\hat{\bm{\zeta}}$ to form the denoised ROI $\hat{\bm{R}}$.
\State Extract the centered $L\times L$ rROI from $\hat{\bm{R}}$.
\State Use the values of the rROI to set the values of the corresponding pixels in $\hat{\bm{I}}$.
\State Rearrange the elements of  $\hat{\bm{u}}$ to form the outliers' ROI.
\State Extract the centered $L\times L$ values of the outliers' ROI.
\State Use these values to set the values of $\hat{\bm{O}}$.
\State Move to the next ROI.
\EndFor
\State Remove the initial padding on $\hat{\bm{I}}$ and $\hat{\bm{O}}$ (if needed).
\end{algorithmic}
\end{algorithm}


\subsubsection{KGARD combined with BM3D (KG-BM3D)}
This is a two-step procedure, that combines the outliers detection properties of KGARD with the denoising capabilities of a standard off-the-shelf denoising method. In this setting (which is the one we propose), the KGARD is actually used to detect the outliers and remove them, while the BM3D wavelet-based denoising method \cite{DaFoKatEg} takes over afterwards to cope with the bounded noise.
Hence, the KGARD algorithm is firstly applied onto the noisy image, to obtain the positions and values of the reconstructed outliers, which are then subtracted from the original noisy image and BM3D is applied to the result.
This method requires the same parameters as KGARD, plus the parameter $s$, which is needed by the BM3D algorithm\footnote{BM3D is built upon the assumption that the image is corrupted by Gaussian noise. Hence, the parameter $s$ is the variance of that Gaussian noise, if this is known a-priori, or some user-defined estimate. However, it has been demonstrated that BM3D can also efficiently remove other types of noise, if $s$ is adjusted properly \cite{bouboulis2010adaptive}.}.

\subsection{Parameter Selection}\label{SEC:par_sel}

This section is devoted on providing guidelines for the selection of the user-defined parameters for the proposed denoising algorithms.  Typical values of $N$ range between 8 and 16. Values of $N$ near 8, or even lower, increase the time required to complete the denoising process with no significant improvements in most cases. However, if the image contains a lot of ``fine details'' this may be advisable. In these cases, smaller values for the width of the Gaussian kernel, $\sigma$, may also enhance the performance, since in this case the regression task is more robust to abrupt changes. However, we should note that $\sigma$ is inversely associated with the size\footnote{For example, if $N=12$ and $\sigma=0.3$, then the kernel width is equal to 3.6 pixels. It is straightforward to see that, if $N$ decreases to say 8, then the kernel width that will provide a length of 3.6 pixels is $\sigma=0.45$.} of the ROI, $N$, hence if one increases $N$, one should decrease $\sigma$ proportionally, i.e., keeping the product $N\cdot\sigma$ constant. We have observed that the values $N=12$ and $\sigma=0.3$ (which result to a product equal to $N\cdot\sigma=3.6$) are adequate to remove moderate noise from a typical image. In cases where the image is rich in details and edges, $N$ and $\sigma$ should be adjusted to provide a lower product (e.g., $N=12$ and $\sigma=0.15$, so that $N\cdot\sigma=1.8$). For images corrupted by high noise, this product should become larger. Finally, $\lambda$ controls the importance of regularization on the final result. Large values imply a strong smoothing operation, while small values (close to zero) reduce the effect of regularization leading to a better fit; however, it may lead to overfitting.

For the experiments presented in this paper, we fixed the size of the ROIs using $N=12$ and $L=8$. These are reasonable choices that provide fast results with high reconstruction accuracy. Hence, only the values for $\sigma$ and $\lambda_0$ need to be adjusted according to the density of the details in the image and the amount of noise. We have found that the values of $\sigma$ that provide adequate results range between 0.1 and 0.4. Similarly, typical values of $\lambda_0$ range from 0.1 to 1. Finally, the constant $E_0$ was set equal to $40$ for all cases.

The parameter $s$ of the BM3D method is adjusted according to the amount of noise presented in the image. It ranges between very small values (e.g, 5), when only a small amount of bounded noise is present, to significantly larger values (e.g., 20 or 40) if the image is highly corrupted.

\subsection{Experiments on Images Corrupted by Synthetic Noise}

\begin{figure}
\centering
\begin{tabular}{cc}
\includegraphics[scale=0.5]{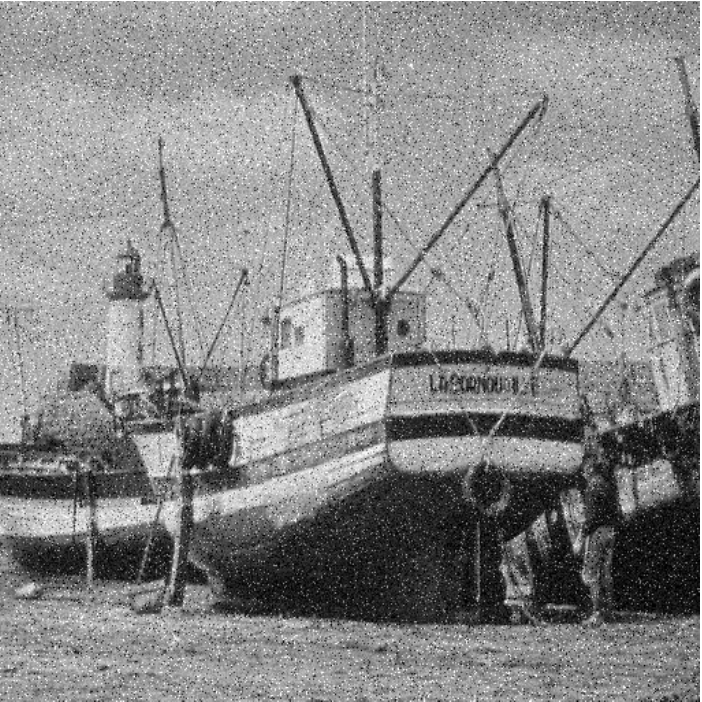} & \includegraphics[scale=0.5]{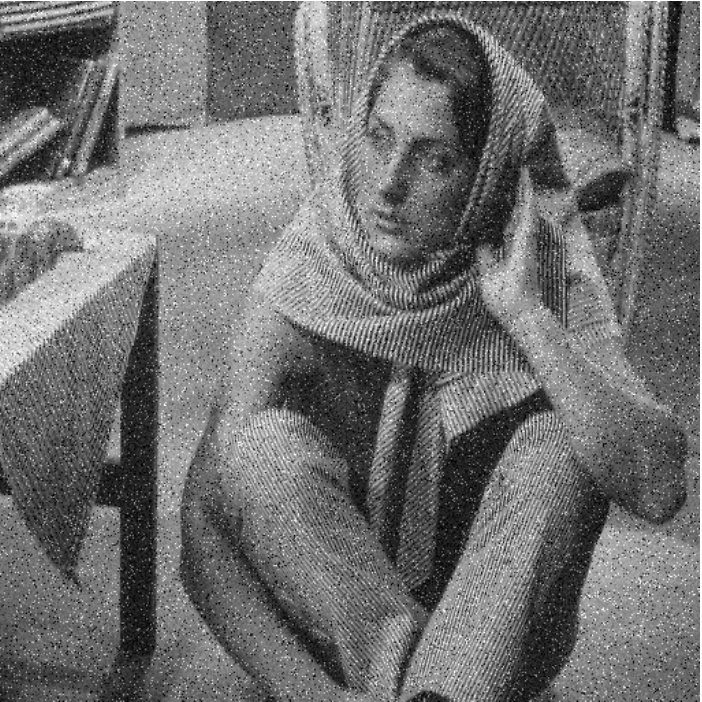}\\
(a) & (b)\\
\includegraphics[scale=0.5]{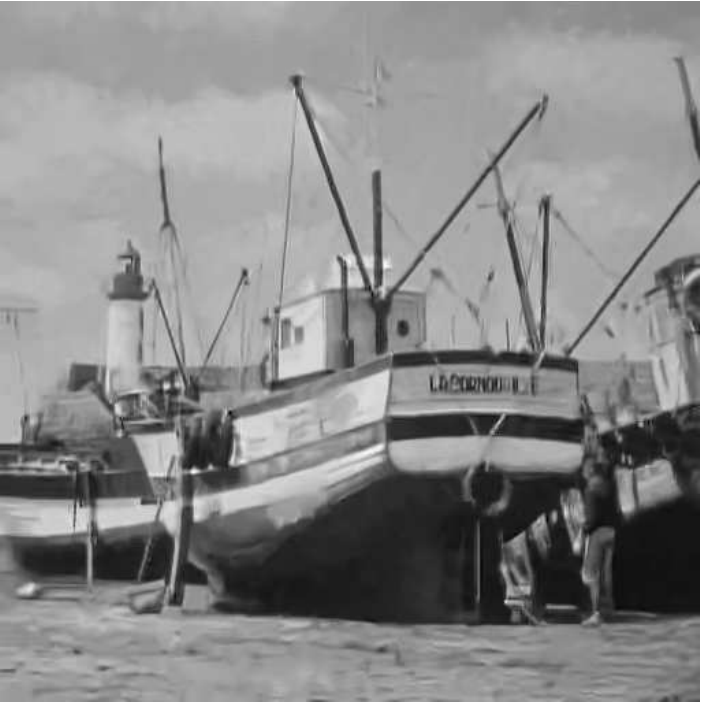} & \includegraphics[scale=0.5]{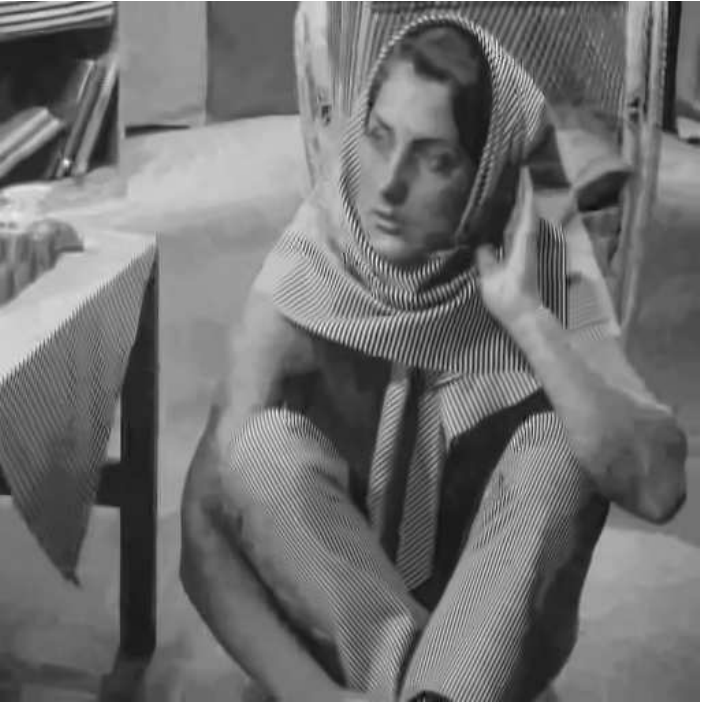}\\
(c) & (d)\\
\includegraphics[scale=0.5]{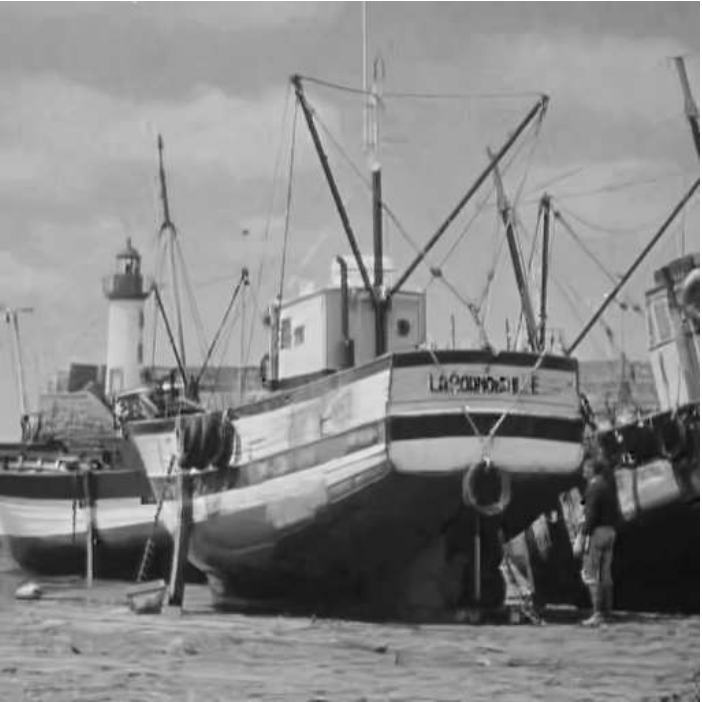} & \includegraphics[scale=0.5]{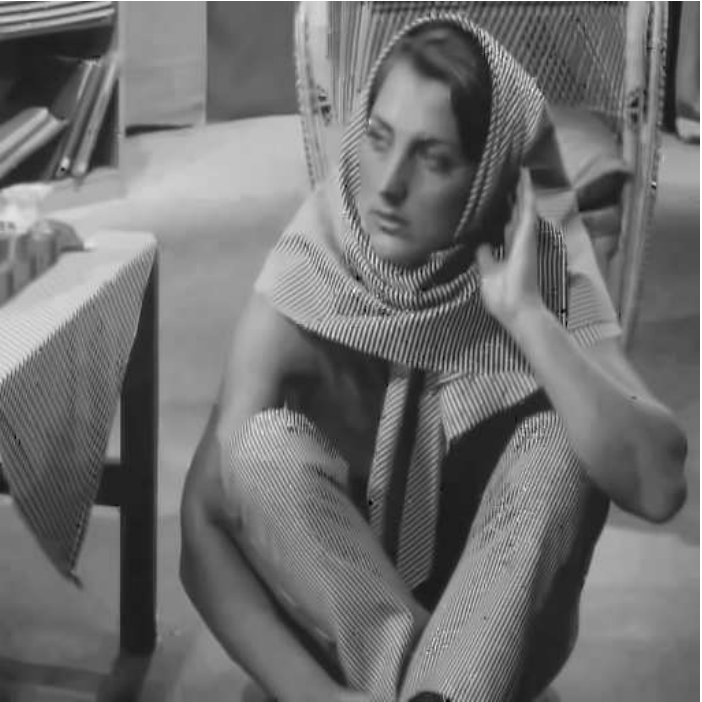} \\
(e) & (f)\\
\end{tabular}
\caption{(a),(b) The \textit{boat} and \textit{Barbara} images corrupted by 20 dB of Gaussian noise and 10\% outliers.
(c), (d) Denoising with BM3D (28.97 dB and 29.2 dB). (e), (f) Denoising with joint KG-BM3D (31.52 dB and 30.43 dB).}\label{FIG:boat_barbara}
\end{figure}

In this section, we present a set of experiments on grayscale images that have been corrupted by mixed noise, which comprises a Gaussian component and a set of impulses ($\pm 100$). The intensity of the Gaussian noise has been ranged between 15 dB and 25 dB and the percentage of impulses from 5\% to 20\%. The tests were performed on three very popular images: the \textit{Lena}, the \textit{boat} and the \textit{Barbara} images, that are included in Waterloo's image repository. Each test has been performed 50 times and the respective mean PSNRs are reported. The parameters have been tuned so that to provide the best result (in terms of MSE). In Table \ref{TAB:denoi1}, the two proposed methods are applied to the \textit{Lena} image and they are compared with BM3D (a state of the art denoising method) and an image denoising method based on (RB-RVM) (``G. N." stands for Gaussian Noise and ``Imp." for Impulses). For the latter, we chose a simple implementation, similar to the one we propose in our methods: the image is divided into ROIs and the RB-RVM algorithm is applied to each ROI sequentially. The parameters were selected to provide the best possible results in terms of PSNR. The size of the ROIs has been set to $N=12$ and $L=8$ for the Lena and \textit{boat} image. As the \textit{Barbara} image has more finer details (e.g., the stripes of the pants) we have set $N=12$ and $L=4$ for this image. Moreover, one can observe that for this image, we have used a lower value for $\sigma$ and $\lambda$ as indicated in Section \ref{SEC:par_sel}. Figure \ref{FIG:boat_barbara} demonstrates the obtained denoised images on a specific experiment (20 dB Gaussian noise and 10\% outliers). It is clear that the proposed method (KG-BM3D) enhances significantly the denoising capabilities of BM3D, especially for low and moderate intensities of the Gaussian noise. If the Gaussian component becomes prominent (e.g., at 15 dB) then the two methods provide similar results. Regarding the computational load, it only takes a few seconds in a standard PC for each one of the two methods to complete the denoising process.

Finally, it is noted that we chose not to include RAM or any $\ell_1$-based denoising method, as this would require efficient techniques to adaptively control its parameters, i.e., $\lambda$, $\mu$ at each ROI (similar to the case of KGARD), which remains an open issue. Having to play with both parameters, makes the tuning computationally demanding. This is because the number of iterations for the method to converge to a reasonable solution increases substantially, once the parameters are moved away from their optimal (in terms of MSE) values\footnote{If the parameters are not optimally tuned, the denoising process may take more than an hour to complete in MATLAB on a moderate computer.}. Finally, it is worth noting that we experimented with noise originating from long-tailed distributions. The results were similar to the ones presented for the classical case of salt and pepper noise.

\begin{table}[t]
\renewcommand{\arraystretch}{0.5}
{\begin{tabular}{|c|c||c|c|c|} \hline
Method & Parameters & G. N. & Imp. & PSNR\\ \thickhline
\textbf{BM3D}   &  $s=30$  & 25 dB  &  5\%  & 32.2 dB \\\hline
\textbf{RB-RVM}    &  $\sigma=0.3$  & 25 dB  &  5\%  & 31.78 dB \\\hline
\textbf{KGARD}   &  $\sigma=0.3$, $\lambda=1$  & 25 dB  &  5\%  &  33.91 dB\\\hline
\textbf{KG-BM3D}   &  $\sigma=0.3$, $\lambda=1$, $s=10$ & 25 dB  &  5\% & \textbf{36.2 dB} \\ \thickhline

\textbf{BM3D}  &  $s=30$  & 25 dB  &  10\%  & 30.84 dB \\\hline
\textbf{RB-RVM}    &  $\sigma=0.3$  & 25 dB  &  10\%  & 31.25 dB \\\hline
\textbf{KGARD}   &  $\sigma=0.3$, $\lambda=1$, $\epsilon=40$  & 25 dB  &  10\%  & 33.49 dB \\ \hline
\textbf{KG-BM3D}   &  $\sigma=0.3$, $\lambda=1$, $s=10$  & 25 dB  &  10\% & \textbf{35.67 dB} \\ \thickhline


\textbf{BM3D}   &  $s=30$  & 20 dB  &  5\%  & 31.83 dB \\\hline
\textbf{RB-RVM}   &  $\sigma = 0.4$  & 20 dB  &  5\%  & 29.3 dB \\\hline
\textbf{KGARD}   &  $\sigma=0.3$, $\lambda=1$  & 20 dB  &  5\%  & 32.35 dB \\\hline
\textbf{KG-BM3D}    &  $\sigma=0.3$, $\lambda=1$, $s=15$ & 20 dB  &  5\% & \textbf{34.24 dB} \\ \thickhline

\textbf{BM3D}  &  $s=35$  & 20 dB  &  10\%  & 30.66 dB \\\hline
\textbf{RB-RVM}   &  $\sigma = 0.4$  & 20 dB  &  10\%  & 29.09 dB \\\hline
\textbf{KGARD}   &  $\sigma=0.3$, $\lambda=1$  & 20 dB  &  10\%  & 31.94 dB \\\hline
\textbf{KG-BM3D}    & $\sigma=0.3$, $\lambda=1$, $s=15$  & 20 dB  &  10\% & \textbf{33.81 dB} \\\thickhline


\textbf{BM3D}   &  $s=35$  & 15 dB  &  5\%  & 30.87 dB \\ \hline
\textbf{RB-RVM}   &  $\sigma=0.6$  & 15 dB  &  5\%  & 26.74 dB \\ \hline
\textbf{KGARD}    &  $\sigma=0.3$, $\lambda=1.5$  & 15 dB  &  5\%  & 29.12 dB \\\hline
\textbf{KG-BM3D}    & $\sigma=0.3$, $\lambda=1$, $s=25$  & 15 dB  &  5\% & \textbf{31.18 dB} \\ \thickhline

\textbf{BM3D}  &  $s=40$  & 15 dB  &  10\%  & 29.94 dB \\\hline
\textbf{RB-RVM}   &  $\sigma=0.4$  & 15 dB  &  10\%  & 25.85 dB \\\hline
\textbf{KGARD}    &  $\sigma=0.3$, $\lambda=2$  & 15 dB  &  10\%  &  28.47 dB  \\\hline
\textbf{KG-BM3D}   &  $\sigma=0.3$, $\lambda=1$, $s=25$  & 15 dB  &  10\% & \textbf{30.77 dB} \\ \thickhline

\end{tabular}
}
\caption{Denoising performed on the \textit{Lena} image corrupted by various types and intensities of noise using the proposed methods, the RB-RVM approach and the state of the art wavelet method BM3D.}\label{TAB:denoi1}
\end{table}

\section{Conclusions}
\label{sec:concl}

In this manuscript, a novel method for the task of non-linear regression in the presence of outliers is presented. The non-linear function is assumed to lie in an RKHS. The proposed scheme iteratively estimates the outliers via a modification of the basic greedy algorithm, i.e., the OMP (or GARD in \cite{papageorgiou2014robust}), while simultaneously estimates the non-linear function via the minimization of a regularized squared error term. Moreover, further improvements and efficient implementations have been established for the method. Theoretical analysis regarding the identification of the outliers in the absence of inlier noise is also provided, results that are absent in other related works. The so-called KGARD is directly compared to state-of-the-art methods and the results demonstrate enhanced performance in terms of in terms of the mean-square-error (MSE) as well as the identification of the outliers. Finally, the method is applied to the task of robust image denoising, for which a parameter-free variant has been established. The proposed method manages to successfully identify the majority of the outliers, which have been modeled as salt and pepper noise (white and black pixels). Furthermore, if combined with the wavelet-based method BM3D, it demonstrates significant gains in terms of the peak-signal-to-noise-ratio (PSNR).


\appendix
\section*{Appendix A. Proof of Theorem \ref{theor:sing_val_bound}}
\label{app:theorem_support}

\begin{proof}
Our analysis is based on the \emph{singular value decomposition} (SVD) for matrix $\bm{X}_{(0)}$ (see Section \ref{subsec:rob_enhcancement}). 

The proposed method attempts to solve at each step, the regularized Least Squares (LS) task \eqref{eq:J_opt} for the selection of matrix $\bm{B}$. The latter task is equivalent to a LS problem in the augmented space\footnote{This is due to the fact that $\bm{B}$ is a projection matrix (based on the $\ell_2$ regularization model).} at each $k$-step, i.e., in \eqref{eq:J_opt2},
where $\bm{X}_{(k)}=\begin{bmatrix}
\bm{X}_{(k-1)} & \bm{e}_{j_k}
\end{bmatrix}$ and $\bm{B}_{(k)}=\begin{bmatrix}
\bm{B}_{(k-1)} & \bm{0} \\ \bm{0}^T & 0
\end{bmatrix}$. Thus, the LS solution at each $k$-step could be expressed as:
\begin{equation}
\label{eq:zksolution}
\hat{\bm{z}}_{(k)}= (\bm{X}_{(k)}^T\bm{X}_{(k)}+\lambda \bm{B}_{(k)})^{-1}\bm{X}_{(k)}^T\bm{y}
\end{equation}
and the respective residual is
\begin{equation}
\label{eq:rkresidual}
\bm{r}_{(k)} =\bm{y} - \bm{X}_{(k)} \hat{\bm{z}}_{(k)}=\bm{y} - \bm{X}_{(k)}(\bm{X}_{(k)}^T\bm{X}_{(k)}+\lambda \bm{B}_{(k)})^{-1}\bm{X}_{(k)}^T\bm{y}.
\end{equation}

\noindent\textbf{Step $k=0$}:\\
Initially, $\bm{B}_{(0)}=\bm{I}_{N+1}$ and $\mathcal{S}_0=  \{1,\dots,N+1\}$ (no index has been selected for the outlier estimate), thus $\bm{X}_{(0)} = [\bm{K}\ \bm{1}]$. Hence, the expression for the initial LS solution $\hat{\bm{z}}_{(0)}$ is obtained from equation \eqref{eq:rkresidual} for $k=0$. By employing the SVD decomposition for matrix $\bm{X}_{(0)}$ in \eqref{eq:Lambda_matrix} and 
combining with \eqref{eq:rkresidual} for $k=0$, we obtain
\begin{equation}
\label{eq:r0_nosubstitution}
\bm{r}_{(0)}= \bm{y} - \bm{QGQ}^T\bm{y},
\end{equation}
where $\bm{G}$ is given in \eqref{eq:Gmatrix}.
Furthermore, substituting $\bm{y}=\bm{X}_{(0)} \orutb+ \orub$ in \eqref{eq:r0_nosubstitution} leads to
\begin{equation}
\label{eq:r0_substitution}
\bm{r}_{(0)} = \orub  + \bm{QFV}^T \orutb - \bm{QGQ}^T \orub ,
\end{equation}
where $\bm{F}=\bm{S}-\bm{GS}=[ \underbrace{\bm{\Sigma} - \bm{G \Sigma}}_{\bm{\Phi}}\ \bm{0} ]$. Matrix $\bm{\Phi}$ is also diagonal, with values
$\phi_{ii}=\frac{\lambda \sigma_i}{\sigma_i^2+\lambda},\  i=1,2,...,N.$
At this point it is required to explore some of the unique properties of matrices $\bm{G}$ and $\bm{F}$. Recall that the (matrix) 2-norm of a diagonal matrix is equal to the maximum absolute value of the diagonal entries. Hence, it is clear that
\begin{equation}
||\bm{G}||_2=\sigma_M^2/(\sigma_M^2+\lambda)\ \text{and}\ ||\bm{F}||_2=||\bm{\Phi}||_2 \leq \sqrt{\lambda}/2,
\label{eq:G_F_bounds}
\end{equation}
since $g(\sigma) = \frac{\sigma^2}{\sigma^2+\lambda}$ is a strictly increasing function of $\sigma \geq 0$ and $\phi(\sigma) = \frac{\lambda \sigma}{\sigma^2+\lambda}$ receives a unique maximum, which determines the upper bound for the matrix 2-norm.

Finally, it should be noted that if no outliers exist in the noise,the algorithm terminates due to the fact that the norm of the initial residual is less than (or equal to) $\epsilon$. However, this scenario is rather insignificant since no robust modeling is required. Thus, if our goal is for the method to be able to handle various types of noise that includes outliers (e.g. Gaussian noise plus impulses), we assume that $\|\bm{r}_{(0)}\|_2 > \epsilon$. In such a case KGARD identifies an outlier selecting an index from the set $\mathcal{\tilde S}_{0}^c=\{1,2,...,N \}$.

At the first selection step, as well as at every next step, we should impose a condition so that the method identifies and selects an index that belongs to the support of the sparse outlier vector. To this end, let $\mathcal{T}$ denote the support of the sparse outlier vector $\orub$. In order for KGARD to select a column $\bm{e}_i$ from matrix $\bm{I}_N$ that belongs to $\mathcal{T}$, we should impose
\begin{equation}
\label{eq:sup_recoveryr0}
|r_{(0),i}|>|r_{(0),j}|,\ \text{for all}\ i \in \mathcal{T}\ \text{and}\ j \in \mathcal{T}^c.
\end{equation}
The key is to establish appropriate bounds, which guarantee the selection of a correct index that belongs to $\mathcal{T}$. Therefore, we first need to develop bounds on the following inner products. Using \eqref{eq:G_F_bounds}, the Cauchy-Schwarz inequality and the fact that $\bm{Q},\bm{V}$ are orthonormal, it is easy to verify that
\begin{flalign}
|\langle \bm{e}_l, \bm{QFV}^T\orutb \rangle |  &\leq \frac{\sqrt{\lambda}}{2} \left\| \orutb \right\|_2 \label{eq:inner_prod_boundF}&\\
\text{as well as  }\  |\langle \bm{e}_l, \bm{QGQ}^T\orub \rangle | &\leq \frac{\sigma_M^2}{\sigma_M^2+\lambda} \left\| \orub \right\|_2,& \label{eq:inner_prod_boundG}
\end{flalign}
for all $l=1,2,...,N$. Thus, for any $i \in \mathcal{T}$, we have that
\begin{flalign}
 && |r_{(0),i}| >& \min{|\oru|} - \frac{\sqrt{2\lambda}}{2} \left\| \orutb \right\|_2 -  \frac{\sigma_M^2}{\sigma_M^2+\lambda }\left\| \orub\right\|_2,
\label{eq:proofofsupp_i}&\\
\textrm{and}\ &&  |r_{(0),j}|  &< \frac{\sqrt{2\lambda}}{2} \left\| \orutb \right\|_2 +  \frac{\sigma_M^2}{\sigma_M^2+\lambda}\left\| \orub\right\|_2,
\label{eq:proofofsupp_j}&
\end{flalign}
for all $j \in \mathcal{T}^c$, where equation \eqref{eq:r0_substitution} and inequalities \eqref{eq:inner_prod_boundF} and \eqref{eq:inner_prod_boundG} have also been used. Hence, imposing \eqref{eq:sup_recoveryr0} leads to \eqref{eq:sigular_value_bound}. It should be noted that, a reason that could lead to the violation of \eqref{eq:sigular_value_bound} is for the term $\min{|\oru|}- \sqrt{2\lambda } \left\| \orutb \right\|_2$ to be non-positive. Thus, since the regularization parameter is fine-tuned by the user, we should select $ \lambda < \left( \min{|\oru|} / \left\| \orutb \right\|_2 \right)^2/2 $. If the condition is guaranteed, then at the first selection step, a column indexed $j_1 \in \mathcal{T}$ is selected. The set of active columns that participates in the LS solution of the current step is then $\mathcal{S}_1=\{ j_1 \} \subseteq \mathcal{T}$ and thus $\bm{X}_{(1)}= \begin{bmatrix}
 \bm{X}_{(0)}  & \bm{e}_{j_1}
\end{bmatrix}$ and $\bm{B}_{(1)}=\begin{bmatrix}
\bm{I}_{N+1} & \bm{0}\\ \bm{0}^T & 0
\end{bmatrix}$.

\noindent\textbf{General $k$ step}:\\
At the $k$ step, $\mathcal{S}_k=\{ j_1,j_2,...,j_k \} \subset \mathcal{T}$ and thus $\bm{X}_{(k)}= \begin{bmatrix}
 \bm{X}_{(0)}  & \bm{I}_{\mathcal{S}_k}
\end{bmatrix}$ and $\bm{B}_{(k)}=\begin{bmatrix}
\bm{I}_{N+1} & \bm{O}_{(N+1)\times k}\\ \bm{O}_{(N+1)\times k}^T & \bm{O}_k
\end{bmatrix}$.
After the selection of the first column, the LS step requires the inversion of the matrix
\[
\small
\bm{D}_{(k)}^T \bm{D}_{(k)} = \begin{bmatrix}
\bm{X}_{(0)}^T \bm{X}_{(0)}+\lambda \bm{I}_{N+1} & \bm{X}_{(0)}^T \bm{I}_{\mathcal{S}_k}\\
\bm{I}_{\mathcal{S}_k}^T \bm{X}_{(0)} & \bm{I}_k		\end{bmatrix}.
\]
By applying the \emph{Matrix inversion Lemma} to $\bm{D}_{(k)}^T \bm{D}_{(k)}$ combined with \eqref{eq:Lambda_matrix} and then substituting into \eqref{eq:rkresidual} leads to:
\begin{equation}
\bm{r}_{(k)} = \bm{P}_{(k)} \orub + \bm{P}_{(k)} \bm{QFV}^T\orutb -\bm{P}_{(k)} \bm{QGQ}^T \orub,
\label{eq:r_k}
\end{equation}
where $\bm{P}_{(k)} = \bm{I}_N + \bm{QGQ}^T \bm{I}_{\mathcal{S}_k} \bm{W}_{(k)}^{-1} \bm{I}_{\mathcal{S}_k}^T - \bm{I}_{\mathcal{S}_k} \bm{W}_{(k)}^{-1} \bm{I}_{\mathcal{S}_k}^T$ and $\bm{W}_{(k)} = \bm{I}_k - \bm{I}_{\mathcal{S}_k}^T \bm{QGQ}^T \bm{I}_{\mathcal{S}_k}$.
If we wish for the algorithm to select an index from the set $\mathcal{T}$, we should impose
$|r_{(k),i}|>|r_{(k),j}|$, for all $i \in \mathcal{T}/\mathcal{S}_k,  j \in \mathcal{T}^c.$
Now $\bm{P}_{(k)}( \orub-\bm{QGQ}^T\orub)= \bm{u}_{(k)}-\bm{QGQ}^T\bm{u}_{(k)},$
where $\bm{u}_{(k)} = \orub_{{\mathcal{T}/\mathcal{S}_k}} +\bm{I}_{\mathcal{S}_k}\bm{W}_{(k)}^{-1} \bm{I}_{\mathcal{S}_k}^T \bm{QGQ}^T\orub_{\mathcal{T}/\mathcal{S}_k}  $. 
Hence, the final form of the residual is:
\begin{equation}
\label{eq:rk_final}
\bm{r}_{(k)} = \bm{u}_{(k)} + \bm{P}_{(k)} \bm{QFV}^T\orutb -\bm{QGQ}^T\bm{u}_{(k)}.
\end{equation}
For $l \notin \mathcal{S}_k$, we conclude that
\[
\bm{P}_{(k)}^T\bm{e}_{l}=\bm{e}_l+ \bm{I}_{\mathcal{S}_k} \bm{W}_{(k)}^{-1} \bm{I}_{\mathcal{S}_k}^T \bm{QGQ}^T\bm{e}_l,
\]
is a $(k+1)$-sparse vector. Furthermore, it is readily seen that,
\begin{equation}
\left\| \bm{W}_{(k)}^{-1} \bm{I}_{\mathcal{S}_k}^T \bm{QGQ}^T\bm{e}_l \right\|_2  \leq \frac{\sigma_M^2}{\lambda}<1,
\label{eq:pkl_bound}
\end{equation}
which leads to
$\left\|  \bm{P}_{(k)}^T\bm{e}_{l} \right\|_2 < \sqrt{2}$. Moreover,
\begin{equation}
|\langle \bm{e}_l, \bm{P}_{(k)} \bm{QFV}^T \orutb \rangle | < \frac{\sqrt{2\lambda}}{2} \left\| \orutb \right\|_2\ \text{and}\ \| \bm{u}_{(k)} \|_2 < \| \orub \|_2.
\label{eq:cond_rk}
\end{equation}
Accordingly, the bounds for the residual are now expressed as
\begin{equation}
\small
|r_{(k),i}| > \min{|\oru|} - \frac{\sqrt{2\lambda}}{2} \left\| \orutb\right\|_2 -  \frac{\sigma_M^2}{\sigma_M^2+\lambda}\left\| \orub \right\|_2,
\label{eq:proofofsupp_irk}
\end{equation}
for any $i \in \mathcal{T}/\mathcal{S}_k$, and
\begin{equation}
|r_{(k),j}|  < \frac{\sqrt{2\lambda}}{2} \left\| \orutb\right\|_2 +  \frac{\sigma_M^2}{\sigma_M^2+\lambda}\left\| \orub\right\|_2,
\label{eq:proofofsupp_jrk}
\end{equation}
for all $j \in \mathcal{T}^c$, where \eqref{eq:rk_final} and \eqref{eq:cond_rk} are used. Finally, imposing the lower bound of \eqref{eq:proofofsupp_irk} to be greater than the upper bound of \eqref{eq:proofofsupp_jrk} leads to the condition \eqref{eq:sigular_value_bound}. At the $k$ step, it has been proved that unless the residual length is below the predefined threshold the algorithm will select another correct atom from the identity matrix and the procedure will repeat until $\mathcal{S}_k=\mathcal{T}.$ At this point, KGARD has correctly identified all possible outliers and it is up to the tuning of the $\epsilon$ parameter whether the procedure terminates (and thus no extra indices are classified as outliers) or it continues and models other extra samples as outliers.
\end{proof}




\bibliographystyle{IEEEtran}
\bibliography{Geo_Library2}

\end{document}